\documentclass{article}

\usepackage{arxiv}

\usepackage[utf8]{inputenc} 
\usepackage[T1]{fontenc}    
\usepackage{hyperref}       
\usepackage{url}            
\usepackage{booktabs}       
\usepackage{amsfonts}       
\usepackage{nicefrac}       
\usepackage{microtype}      
\usepackage{lipsum}		
\usepackage{graphicx}
\usepackage{natbib}
\usepackage{doi}
\usepackage{svg}
\usepackage{float}
\usepackage{booktabs}
\usepackage{multicol}
\usepackage[usestackEOL]{stackengine}
\usepackage{makecell}
\usepackage[labelfont=bf]{caption}
\usepackage{etoolbox}
\AtBeginEnvironment{thebibliography}{\interlinepenalty=10000}
\usepackage{lineno}

\title{A multiscale spatiotemporal approach for smallholder irrigation detection}

\author{
  Terence Conlon\thanks{Corresponding author} \\
  Department of Mechanical Engineering \\
  Columbia University\\
  New York, NY, USA\\
  \texttt{terence.conlon@columbia.edu} \\
   \And
  Christopher Small\\
  Lamont Doherty Earth Observatory \\
  Columbia University\\
  Palisades, NY, USA\\
  \texttt{csmall@columbia.edu} \\
  \And 
  Vijay Modi \\
  Department of Mechanical Engineering \\
  Columbia University\\
  New York, NY, USA\\
  \texttt{modi@columbia.edu} \\
}

\renewcommand{\headerleft}{Conlon et al.}
\renewcommand{\undertitle}{}
\renewcommand{\shorttitle}{}

\begin{document}
\maketitle

\begin{abstract}
In presenting an irrigation detection methodology that leverages multiscale satellite imagery of vegetation abundance, this paper introduces a process to supplement limited ground-collected labels and ensure classifier applicability in an area of interest. Spatiotemporal analysis of MODIS 250m Enhanced Vegetation Index (EVI) timeseries characterizes native vegetation phenologies at regional scale to provide the basis for a continuous phenology map that guides supplementary label collection over irrigated and non-irrigated agriculture. Subsequently, validated dry season greening and senescence cycles observed in 10m Sentinel-2 imagery are used to train a suite of classifiers for automated detection of potential smallholder irrigation. Strategies to improve model robustness are demonstrated, including a method of data augmentation that randomly shifts training samples; and an assessment of classifier types that produce the best performance in withheld target regions. The methodology is applied to detect smallholder irrigation in two states in the Ethiopian highlands, Tigray and Amhara, where detection of irrigated smallholder farm plots is crucial for energy infrastructure planning. Results show that a transformer-based neural network architecture allows for the most robust prediction performance in withheld regions, followed closely by a CatBoost random forest model. Over withheld ground-collection survey labels, the transformer-based model achieves 96.7\% accuracy over non-irrigated samples and 95.9\% accuracy over irrigated samples. Over a larger set of samples independently collected via the introduced method of label supplementation, non-irrigated and irrigated labels are predicted with 98.3\% and 95.5\% accuracy, respectively. The detection model is then deployed over Tigray and Amhara, revealing crop rotation patterns and year-over-year irrigated area change. Predictions suggest that irrigated area in these two states has decreased by approximately 40\% from 2020 to 2021.
\end{abstract}

\keywords{irrigation detection \and spatiotemporal modeling \and multiscale imagery \and machine learning \and Ethiopia}

\section{Introduction}

Between 1970 and 2008, global irrigated area increased from 170 million to 304 million hectares \citep{Vogels2019}. In sub-Saharan Africa however, as little as 4-6\% of cultivated area is irrigated, given the lack of electric grid infrastructure and the high cost of diesel \citep{Wiggins2021}. Locating isolated irrigation identifies areas that can support higher quality energy provision services -- e.g. a grid connection or minigrid installation -- as these sites can sustain higher energy demands and the attendant electricity costs \citep{conlon2020}. Facilitated through informed planning, irrigation expansion has a direct impact on poverty reduction: In Ethiopia, one study found that the average income of irrigating households was double that of non-irrigating households \citep{Gebregziabher2009}. 

In data poor locations, satellite imagery provides a source of detailed synoptic observations of irrigated agriculture \citep{ShahriarPervez2014}. A previous irrigation mapping effort in Ethiopia used three 1.5m resolution SPOT6 images to distinguish between large-scale and smallholder irrigation in the Ethiopian rift \citep{Vogels2019}. This approach was then adapted to intake a timeseries of 10m Sentinel-2 imagery to predict irrigation presence across the horn of Africa \citep{Vogels2019a}. While both studies demonstrated high accuracies over collected observations, limited labels precluded a more rigorous performance assessment over the entire area of interest. Other studies have used multiscale imagery to detect irrigation, including one that fuses MODIS and Landsat imagery to identify irrigated extent, frequency, and timing in northwestern China \citep{Chen2018}. Here, unique advantages of satellite imagery products at different resolutions are exploited: 250m MODIS imagery is valuable for characterizing vegetation over large areas \citep{huete1999modis}, while decameter resolution imagery from Landsat or Sentinel-2 missions can better discern plot extent \citep{Phiri2019}. 

Deep learning techniques have become widely used for land process classification, as they uncover intricate structures in large, complex datasets \citep{Lecun2015}; and provide a robust method of handling phenological variability \citep{Zhong2019}. However, despite increasing availability of remotely sensed imagery, computing resources, and advanced algorithms for information extraction, high-quality labels remain scarce and expensive to acquire. Methods of overcoming label scarcity generally fall into one of four categories: 1) using pretrained networks; 2) unsupervised and self-supervised learning; 3) data augmentation; or 4) additional label collection \citep{Li2018}. Even as pretrained networks like ImageNet \citep{JiaDeng2009} are highly effective for true-color image classification, these networks' weights do not translate to tasks that intake multispectral or hyperspectral imagery \citep{Tao2020}. Unsupervised learning techniques, including those that ensemble different clustering methods -- e.g. \citet{Banerjee2015} -- have been shown to effectively organize unlabeled imagery. Existing work has also demonstrated that training a Generative Adversarial Network (GAN) -- itself a type of unsupervised learning -- has allowed for improved change detection performance on multispectral imagery, e.g. \citet{Saha2019}. For data augmentation, three techniques are often implemented: image translation, rotation, and flipping \citep{Yu2017, Stivaktakis2019}; however, these techniques do not have obvious analogues for pixel-based classification. 

In assessing the impact of training dataset size on land cover classification performance, \citet{Ramezan2021} finds that investigating multiple types of classifiers is recommended, as the performance of specific classifiers is highly dependent on the number of training samples. A number of other studies have introduced methods for obtaining training samples, including collection via hand-engineered rules \citep{Abbasi2015}; normalized difference in vegetation index (NDVI) thresholding \citep{Bazzi2021}; finding neighboring pixels that are highly similar to labeled pixels \citep{Naik2021}; and visual inspection of high-resolution \citep{Vogels2019a} and decameter resolution \citep{Wu2016} imagery. Lastly, while larger training datasets generally yield better model performance, condensing input samples via dimensionality reduction has been demonstrated to increase land cover classification accuracy \citep{Sivaraj2022, Stromann2020}.

Another lingering issue in land process mapping is determining the conditions under which a model can be utilized in locations beyond where it was trained. Site-specific methods may not be easily transferable to other places or climes \citep{Ozdogan2010,Bazzi2020}, and the performance of transferred models can often only be assessed \textit{after} full implementation in a novel setting \citep{DeLima2020}. Therefore, processes that yield insights about model transferability \textit{before} training and inference offer benefits to researchers seeking to understand the maximum spatial applicability of their approaches. 

As current methods primarily focus on already well-understood areas of interest with existing datasets, new techniques and products need to be developed for parts of the world lacking labeled data. In the realm of irrigation detection, new methodologies and mapping products can help identify locations for further energy system planning and investment, as these areas contain latent energy demands that can make higher quality energy services cost-effective and increase incomes. To this end, the following manuscript presents a multiscale methodology that leverages 250m MODIS imagery for regional phenological characterization and 10m Sentinel-2 imagery for irrigation detection on smallholder plots. This approach is then applied to the 205,000 km\textsuperscript{2} Ethiopian highlands, whereby it introduces a novel method of label collection; an evaluation of different classifier architectures and training strategies that ensure model applicability within the area of interest; and an assessment of irrigated area in the Tigray and Amhara states of Ethiopia for 2020 and 2021.

\section{Background}

\begin{figure}[htp]
\begin{center}
\includegraphics[width=\textwidth]{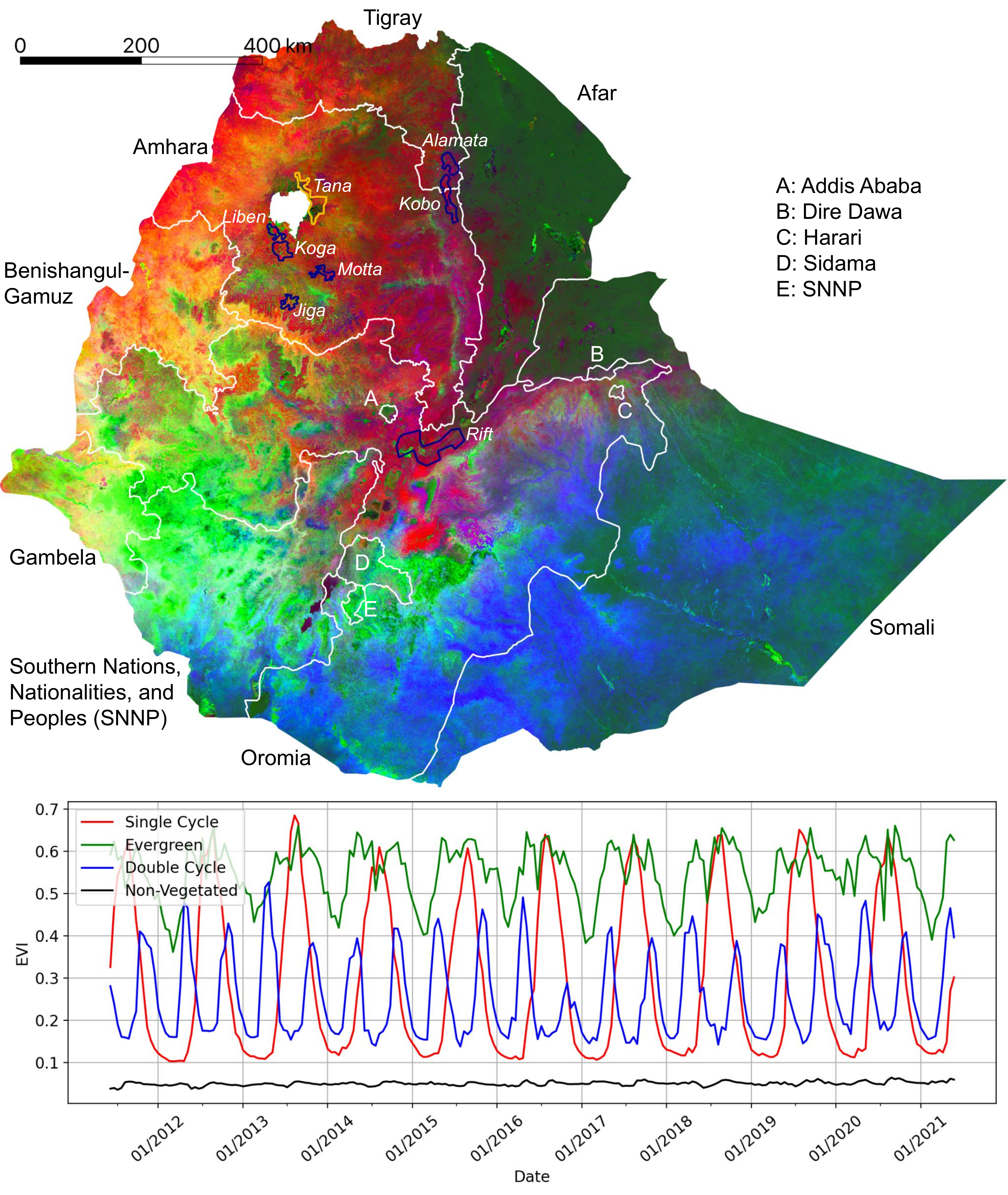}
\end{center}
\caption{Continuous endmember fraction map derived from a temporal mixture model of 250m MODIS enhanced vegetation indices (EVI). Smooth gradients and abrupt transitions in phenology are primarily related to topography and variations in precipitation. Region names showing locations of labeled polygons are italicized: The region containing ground collection (GC) labels is delineated in gold; the regions containing visual collection (VC) labels are delineated in blue.}
\label{tem}
\end{figure}

Identification of dry season greening as potentially irrigated agriculture must take into account spatiotemporal variations in native vegetation phenological cycles. The complex topography of the Ethiopian Highlands and East African rift system, combined with the latitudinal movement of the InterTropical Convergence Zone (ITCZ) and seasonal upwelling of the Somali current in the Arabian Sea produces a diversity of rainfall patterns that control annual vegetation phenological cycles in the study area\footnote{See \citet{Wakjira2021} for a fuller discussion of rainfall patterns in Ethiopia.}. In order to provide phenological context with which to identify anomalous dry season greening, a regional vegetation phenology map is derived from spatiotemporal analysis of timeseries of vegetation abundance maps. Using the spatiotemporal characterization and temporal mixture modeling approach given by \citep{Small2012} applied to timeseries of MODIS enhanced vegetation index (EVI) maps, four temporal endmember (tEM) phenologies are identified that bound the temporal feature space of all vegetation phenology cycles observed on the East African Sahel. These four tEM phenologies form the basis of a linear temporal mixture model that can be inverted to provide tEM fraction estimates for each pixel's vegetation phenology.  Figure \ref{tem} presents a spatiotemporal phenological characterization for the country, created from 16-day 250m MODIS EVI imagery between June 1\textsuperscript{st}, 2011 and June 1\textsuperscript{st}, 2021.

The four tEMs extracted for Ethiopia are as follows: a \textit{single cycle} tEM, representing a single annual vegetation cycle per year that peaks in September/October; an \textit{evergreen} tEM, representing perennial vegetation; a \textit{double cycle} tEM, representing semiannual vegetation cycles observed on the Somali peninsula; and a \textit{non-vegetated} tEM, representing barren or non-existent vegetation. The ensuing phenology map in Figure \ref{tem} contains unmixing root mean square (RMS) error less than 10\% for 90\% of the pixels; additional unmixing error statistics and the locations of the extracted tEMs in principal component (PC) feature space are shown in Supplementary Figures S1-S2. 

Figure \ref{tem} roughly divides into 4 quadrants. In the northeast quadrant, Afar appears as dark green, indicating that none of the 4 tEMs contribute significantly to phenologies in this part of the country: The vegetation that does exist in this mostly barren area is represented by low levels of evergreen tEM abundances. In the southeast quadrant, dominated by Somali and a portion of Oromia, vegetation patterns cycle twice annually. This is an area with bimodal rainfall but low total annual precipitation that results in the \textit{double cycle} tEM containing peak vegetation abundances lower than those of the \textit{single cycle} and \textit{evergreen} tEMs. It follows that southeast Ethiopia is more pastoral with sparser vegetation than other parts of the country. 

The southwest quadrant – covering Southern Nations, Nationalities, and Peoples' (SNNP) Region, Sidama, and the western portion of Oromia – contains significant amounts of evergreen vegetation, as is demonstrated by its bright green hue. Here, evergreen vegetation is supported by bimodal rainfall with higher levels of annual precipitation than in eastern Ethiopia. In contrast, the northwest quadrant of the phenology map contains red-dominant color gradients, indicating phenologies similar to the \textit{single cycle} tEM. This portion of the country, known as the Ethiopian highlands and comprising of Amhara and Tigray, is highly agricultural; the main cropping season lasts from June to October and coincides with the primary \textit{kiremt} rains, with some secondary cropping following the lighter \textit{belg} rains from March to May. Accordingly, cropping that occurs during the dry season between November and March is likely to be irrigated.  

In presenting a map of dominant vegetation phenologies in Ethiopia, Figure \ref{tem} provides a guide for land cover classification applicability within the country. For instance, a dry season irrigation detector trained in Amhara will perform poorly in SNNP, as phenological patterns differ significantly across these states, and dry season crop cycles exhibit different vegetation signatures. In contrast, a dry season irrigation detector developed across Amhara can be transferred to Tigray or Benishangul-Gamuz, due to regional phenological similarities. 

The named, italicized outlines in Figure \ref{tem} represent the 8 areas containing labels used in this manuscript, referred to as \textit{regions}: The yellow outline indicates a region where labels were collected via a ground survey, and the purple outlines indicate regions where labels were collected by means of visual interpretation and timeseries inspection. Full information on the labeled data collection process is presented in Section 3. 

\section{Materials and Methods}

The data collection portion of this manuscript’s methodology consists of pairing Sentinel-2 imagery with labeled polygons to train an irrigation detector. Here, a pixel timeseries paired with a binary irrigation/non-irrigation label constitutes a sample. Irrigation is defined as such: A pixel is irrigated if its phenology includes at least one non-perennial vegetation cycle during the dry season, December 1\textsuperscript{st} to April 1\textsuperscript{st} for the Ethiopian highlands. Conversely, a pixel is non-irrigated if its phenology demonstrates only vegetation growth that can be attributed to the area’s known rainy seasons. Irrigated areas are only of interest if they contain dry season vegetation cycles; this strict definition of irrigation excludes supplemental irrigation practices and perennial crops that may be consistently irrigated throughout the year.

\subsection{Sentinel-2 imagery collection}

The following analysis uses bottom-of-atmosphere corrected (processing level L2A) Sentinel-2 temporal stacks – four dimensional arrays created by stacking a set spatial extent of imagery bands over multiple timesteps – using the Descartes Labs (DL) platform, a commercial environment for planet-scale geospatial analysis. Images are collected at a 10-day time resolution. To focus on the 2020 and 2021 dry seasons, the time period of interest is defined as between June 1\textsuperscript{st}, 2019, and June 1\textsuperscript{st}, 2021. Given the 10-day timestep, 72 image mosaics are collected – 36 per year. Additional information on the imagery download process is available in the Supplementary Materials. 

\subsection{Label collection}

Two types of labeled data are leveraged for irrigation mapping: \textit{ground collection} (GC) labels, acquired via an in-person survey; and \textit{visual collection} (VC) labels, acquired via visual identification of dry season vegetation from Sentinel-2 imagery using the DL platform and subsequent cleaning via timeseries clustering. The locations of these GC and VC regions are shown in italics in Figure \ref{tem}, with all labels collected for the 2021 dry season. A description of the ground collection survey is presented in the Supplementary Materials. As the GC labels constitute our highest quality irrigation observations, verified by in situ visits to individual plots, we do not use them for training during the model sensitivity analysis, instead reserving them for validation of classifier performance. 

\subsubsection{Visual label collection}

To supplement the GC labels located in Tana, visually collected labels are acquired for seven separate regions via a three-step process of 1) visual inspection, 2) EVI timeseries confirmation, and 3) cluster cleaning. Each of these steps is described in its eponymous subsection below. 

\paragraph{Visual inspection}

The first step in the VC labeling process involves drawing polygons around locations that either: a) present as cropland with visible vegetation growth (for the collection of irrigated samples), or b) present as cropland with no visible vegetation growth (for the collection of non-irrigated samples), based on dry-season, false-color Sentinel-2 imagery presented on the DL platform. Sub-meter resolution commercial satellite imagery from Google Earth Pro is also used to confirm the existence of cropland in the viewing window. For the collection of non-irrigated labels, polygons are restricted to areas that contain non-perennial cropland; however, because only phenologies that contain dry season vegetation cycles are considered irrigated, non-irrigated polygons occasionally overlap other types of land cover – e.g., perennial crops, fallow cropland, or areas with human settlement – with any overlap likely to improve training robustness. 

\begin{figure}[ht]
\begin{center}
\includegraphics[width=\textwidth]{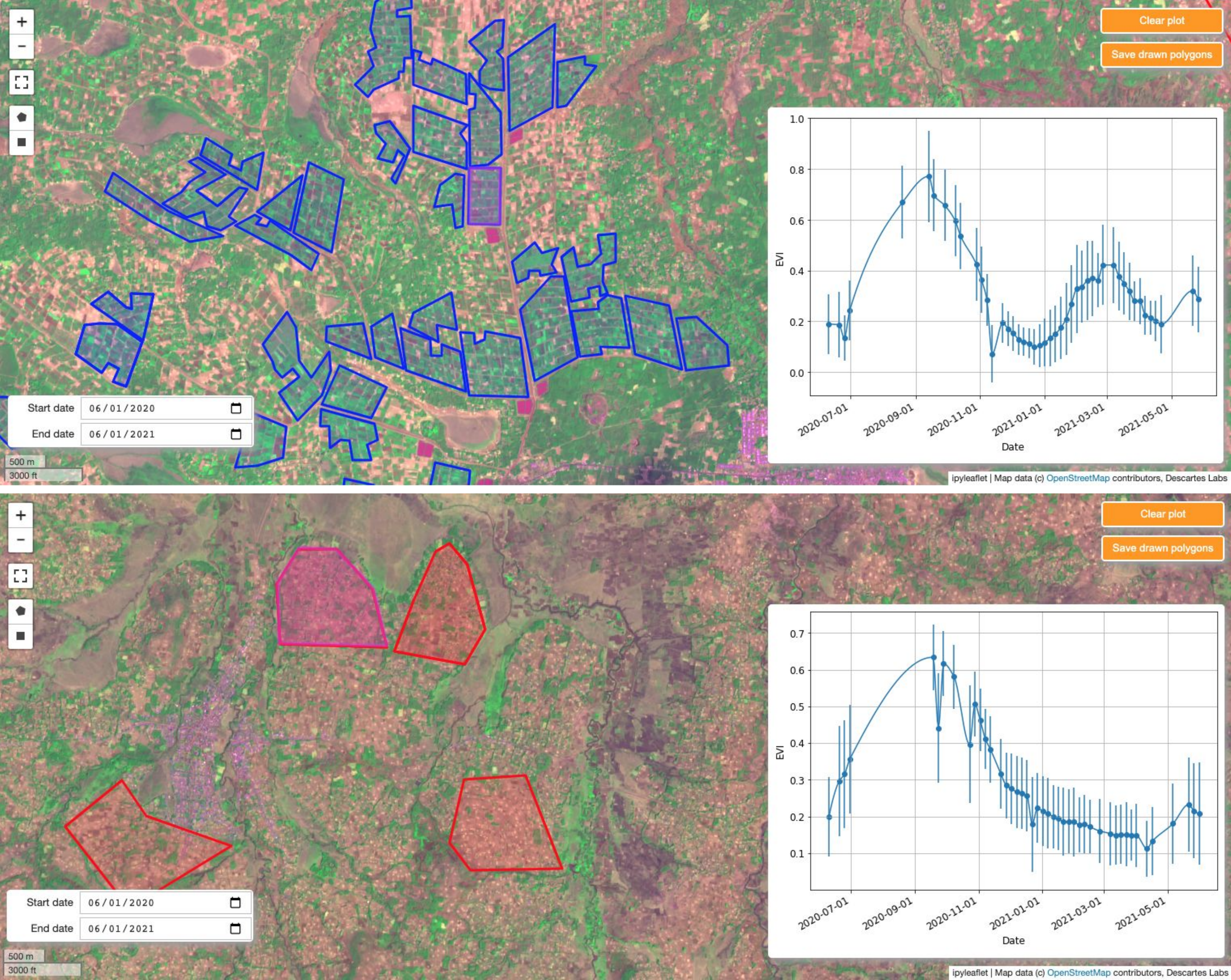}
\end{center}
\caption{Example of the visual collection (VC) labeling process in Koga using the Descartes Labs platform. Blue polygons denote areas determined to be irrigated; red polygons are determined to be non-irrigated. Background imagery is a false-color Sentinel-2 image taken in March 2021: red, near-infrared, and blue bands are presented in the RGB channels, respectively. In (a), the Sentinel-2 enhanced vegetation index (EVI) timeseries is shown for the drawn purple rectangle in the middle of the window; in (b), the Sentinel-2 EVI timeseries is shown for the drawn pink, semi-octagonal polygon in the top left of the window. Both timeseries present the median EVI values for all pixels contained within the drawn polygon; the error bars show one standard deviation of these values above and below the median. In both figures, the drawn polygons are confirmed as VC labels, since they meet the definitions of irrigation/non-irrigation, respectively.}\label{vv_collection}
\end{figure}

\paragraph{EVI timeseries confirmation}
After drawing a polygon around a suspected irrigated or non-irrigated area, the second step in the VC label acquisition process entails inspection of the median Sentinel-2 EVI timeseries of all pixels contained within the polygon; this step is shown in the plot windows of Figure \ref{vv_collection}. Here, all available Sentinel-2 imagery with less than 20\% cloud cover between June 1, 2020, and June 1, 2021 is retrieved; a cubic spline is then fit to all available data to generate continuous EVI timeseries. For potential irrigated polygons, if the EVI timeseries shows a clear peak above 0.2 during the dry season, it is confirmed as irrigated. Similarly, for potential non-irrigated polygons, an EVI timeseries that demonstrates a single vegetation cycle attributable to Ethiopia’s June to September rains is taken as confirmation of a non-irrigated VC polygon. However, if the EVI timeseries does not confirm the expected irrigated/non-irrigated class, or if the plotted EVI error bars (representing $\pm$one standard deviation of the EVI values at that timestep) indicate a level of signal noise within the polygon that prevents the identification of a clear vegetation phenology, the polygon is discarded. 

Figure \ref{vv_collection}(a) demonstrates an example of irrigated VC label collection in the Koga region – here, the double vegetation peak present in the EVI timeseries confirms the purple polygon in the center of the window as irrigated (blue polygons indicate areas already saved as irrigated VC labels). Figure \ref{vv_collection}(b) demonstrates the same process for non-irrigated VC labels, also in Koga: The single EVI peak in October 2020 confirms the pink polygon in the top left of the window as non-irrigated (red polygons indicate areas already saved as non-irrigated VC labels).  

\begin{figure}[ht]
\begin{center}
\includegraphics[width=\textwidth]{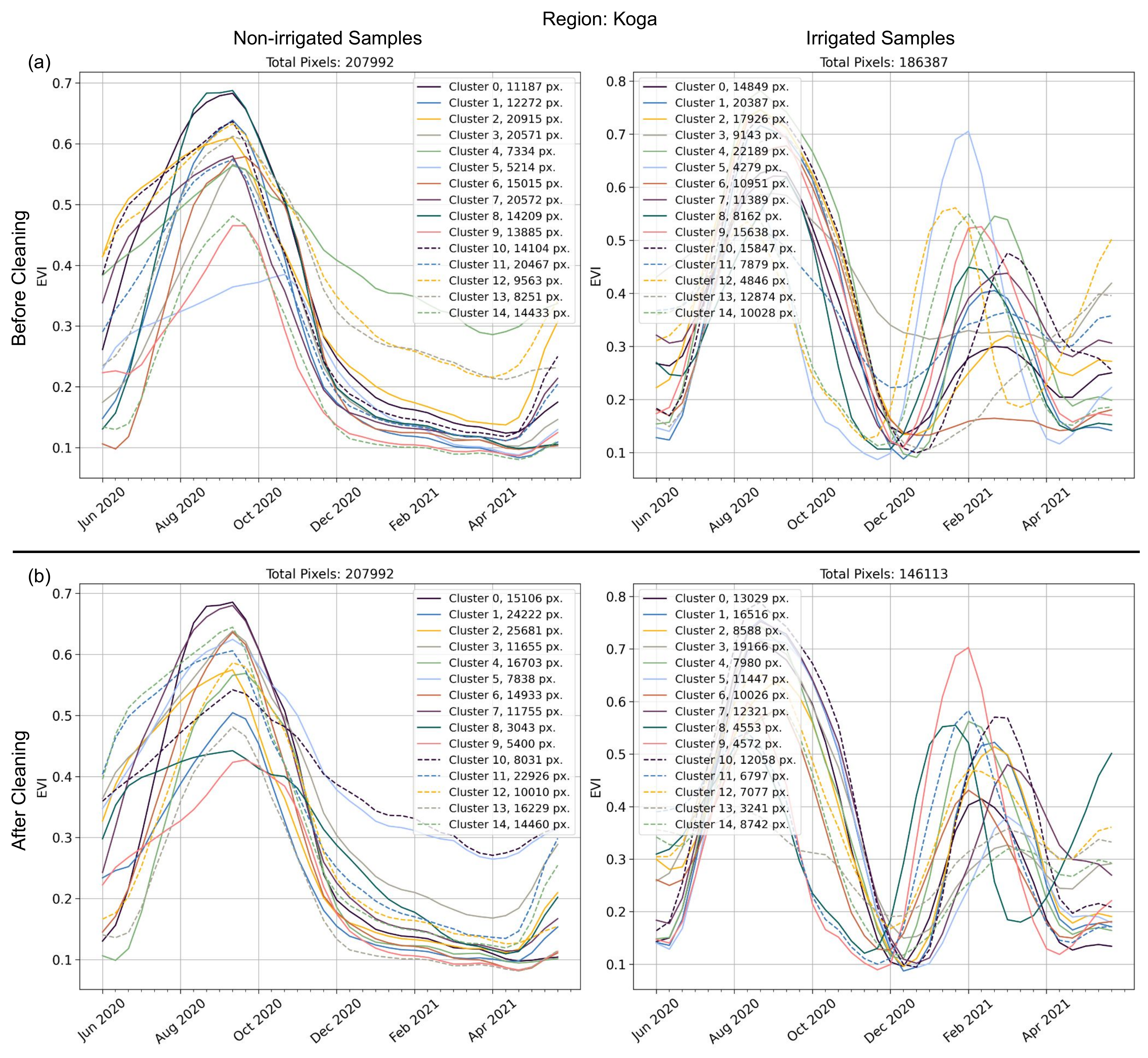}
\end{center}
\caption{Clustered enhanced vegetation index (EVI) timeseries before and after cluster cleaning for the Koga visual collection (VC) region. Before and after cleaning, pixels are grouped into one of 15 randomly indexed clusters. In (a), Clusters 3, 6, and 13 of the irrigated samples are discarded due to either (6, 13) not containing a clear EVI peak above 0.2 during the dry season (December 1\textsuperscript{st} to April 1\textsuperscript{st}); or (3) not containing successive EVI values below 0.2. All non-irrigated clusters display a single vegetation peak aligned with the main rainy season, and the irrigated clusters after cleaning (b) all display a vegetation cycle during the dry season.}\label{cleaning_labels}
\end{figure}

\paragraph{Cluster cleaning}
The third step in the VC label acquisition process involves bulk verification of the collected timeseries by means of cluster cleaning. For each VC region, all pixels that reside within labeled polygons are collected and split based on the irrigated/non-irrigated class labels of the polygons. Fifteen-component Gaussian mixture models are fit to each class’s data to extract the dominant phenologies contained within the region’s samples; the EVI timeseries representing the cluster centroids are then plotted, with the plot legend displaying the number of samples per cluster. Figure \ref{cleaning_labels}(a) presents the results of this initial clustering for the Koga region. 

From the initial cluster timeseries, an iterative process begins to ensure that all cluster timeseries align with the specified class label. For an irrigated cluster timeseries to be kept, it must contain multiple successive EVI values above and below 0.2, and it must contain a clear EVI peak above 0.2 during the dry season. Analogously, non-irrigated cluster timeseries are discarded if they display a clear dry-season EVI peak above 0.2. If these conditions are not met – as is the case for Clusters 3, 6, and 13 of the Koga irrigated samples, which do not contain a clear EVI peak above 0.2 between December 1, 2020 and April 1, 2021 (Clusters 6 and 13) or do not senesce below an EVI threshold of 0.2 for successive timesteps (Cluster 3) – all pixel timeseries associated with that cluster are discarded from the labeled data. This process is repeated until all 15 clusters for both classes demonstrate EVI signals that meet the non-irrigated/irrigated class definitions. The final, cleaned cluster timeseries for the Koga region are shown in Figure \ref{cleaning_labels}(b).  

Cluster-cleaning is performed for all regions’ labeled data, including labeled data collected from the GC region, Tana. For increased visibility into the labeled data collected and used for training, these regions’ clusters before and after cleaning are included in Appendix A of the Supplementary Materials. 

A summary of the number of collected polygons and cleaned pixel timeseries samples in each region is shown in Supplementary Tables S2-S3: In total, 1,207,233 non-irrigated samples and 907,887 irrigated samples are used, taken from 1702 and 750 labeled polygons, respectively. For model training and evaluation, data are divided among training, validation, and test splits\footnote{In splitting the labeled data, the training/validation/testing terminology standard in machine and deep learning literature is adopted.}. Here, polygons in each labeled region are split according to a 70/15/15 training/validation/test ratio; this method ensures that highly similar pixels from within the same polygon do not exist across training configurations, a division of data that would artificially inflate model performance for the task of predicting irrigation over pixel timeseries unseen by the model. All training, validation, and testing is performed pixelwise (i.e., having removed the spatial relationships of samples). 

The Supplementary Materials contain additional information about the labeled data distributions, including a statistical evaluation of the similarity of labeled samples across region and class (see Supplementary Tables S4-S5)

\subsection{Prediction admissibility criteria}

Given that irrigated phenologies exist over a small fraction of the total land area of the Ethiopian highlands, and that there are many types of land cover that do not fall within this manuscript's non-irrigated/irrigated cropland dichotomy, a set of criteria are imposed to exclude pixel phenologies that are not cropland or are highly unlikely to be irrigated. Table \ref{tab:admissibility_criteria} presents five criteria that must all be met for a pixel timeseries to be potentially irrigated and the motivation behind each.

\begin{table}[t]
 \caption{Prediction admissibility criteria. All criteria need to be satisfied for a prediction to be admitted as irrigated.}
  \centering
  \begin{tabular}{ll}
    \toprule
    Admissibility Criteria & Motivation \\ \midrule \midrule
    10\textsuperscript{th} percentile of EVI timeseries $<$ 0.2 & Remove evergreen pixels \\ \midrule
    90\textsuperscript{th} percentile of EVI timeseries $>$ 0.2 & Remove barren/non-vegetated pixels \\ \midrule
    \makecell[cl]{Maximum of the EVI timeseries during \\the dry season (Dec. 1 – Apr. 1) $>$ 0.2} &	 \makecell[cl]{Remove pixels with no vegetation\\growth in the dry season} \\ \midrule
    \makecell[cl]{Ratio of the 90\textsuperscript{th}:10\textsuperscript{th} percentile of the\\EVI timeseries $>$ 2} &  \makecell[tl]{Remove evergreen pixels} \\ \midrule
    \makecell[cl]{Shuttle Radar Topography Mission slope\\ measurement $<$ 8\%}& \makecell[cl]{Remove pixels in highly sloped\\ settings where cropping is impractical}\\

    \bottomrule
  \end{tabular}
  \label{tab:admissibility_criteria}
\end{table}

These vegetation-specific criteria are informed by the EVI distributions of labeled irrigated samples for all label collection regions: Supplementary Figure S3 contains cumulative distribution functions (CDFs) for the 10\textsuperscript{th} and 90\textsuperscript{th} EVI timeseries percentiles, the 90\textsuperscript{th}:10\textsuperscript{th} EVI timeseries percentile ratio, and the maximum EVI value during the dry season. CDFs are presented for all regions’ irrigated samples, including for a set of polygons collected over evergreen land cover areas. 

The criteria in Table \ref{tab:admissibility_criteria} are also used to create a reference irrigation classifier that does not rely on machine learning. For this reference classifier, if all 5 conditions are met, the sample is deemed irrigated; if any of the conditions is not satisfied, the sample is deemed non-irrigated.

\subsection{Model training}

\subsubsection{Model architectures}

\begin{figure}[ht]
\begin{center}
\includegraphics[width=12cm]{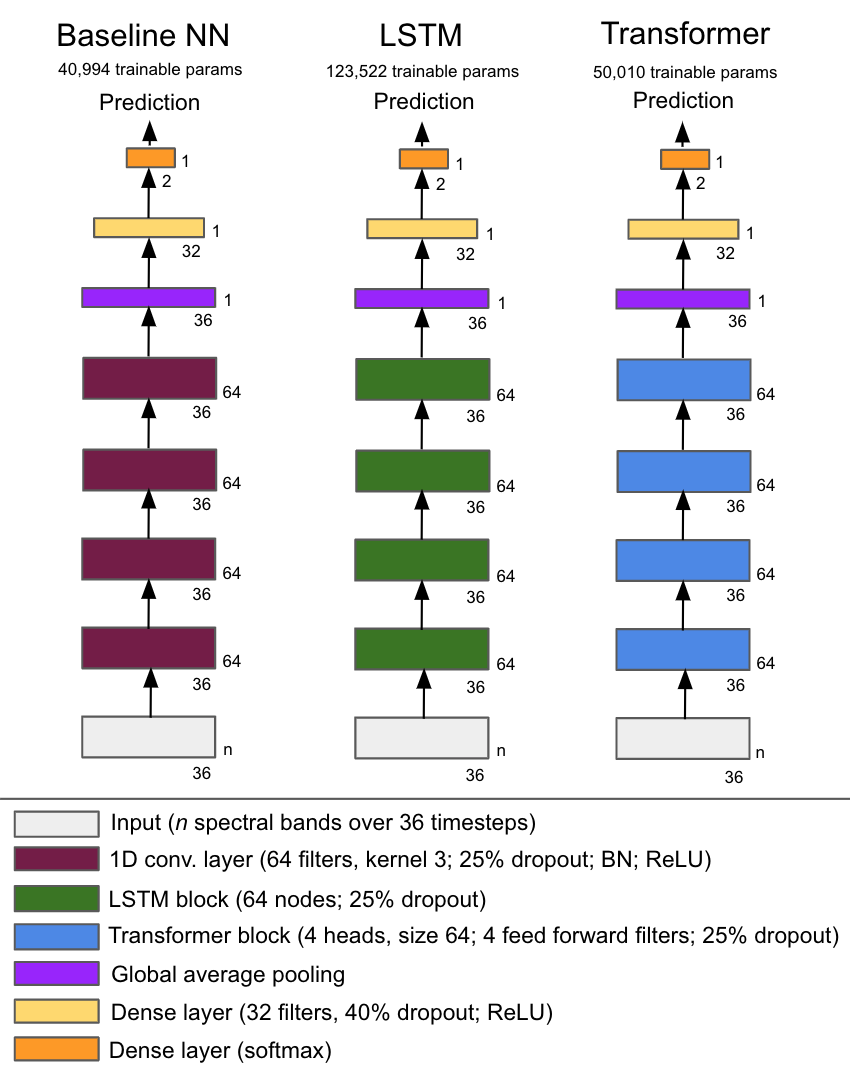}
\end{center}
\caption{Neural network (NN) model architectures tested as irrigation detection classifiers. Model architectures are consistent by design; only encoding blocks differ across networks.}
\label{model_architectures}
\end{figure}

Five separate classifier types are compared to determine the model architecture with the most robust irrigation detection performance across regions. The first two classifiers are decision tree-based: A random forest with 1000 trees \citep{Breiman2001}; and a CatBoost model that uses gradient boosting on up to 1000 trees \citep{prokhorenkova2019catboost}. The other three classifiers are neural networks (NN): A baseline network, an LSTM-based network, and a transformer-based network. For comparability, these three classifier architectures are designed to have similar structures, based on the strong baseline model structure proposed in \citep{Wang2017}; as seen in Figure \ref{model_architectures}, they differ only in the type of encoding blocks used. 

\subsubsection{Model training strategy}

The implemented model training strategy addresses two potential pitfalls among training processes: 1) imbalanced samples across region and class; and 2) high similarity among samples within a region that may not reflect the sample distributions across all regions. Consistent with best practices in dealing with imbalanced data, this first issue is addressed with a) class balancing weights specific to each region, based on the “balanced” heuristic inspired by \citet{King2001}; and b) a region-specific weight equal to the ratio of the maximum number of samples in any region to the number of samples for the region in question. Both class-balancing and region-balancing weights are used in all training configurations. 

To address potential redundancy and time-specificity among samples within a region, random shifts are applied to all input timeseries. The sizes of these random shifts vary between -3 and +3 timesteps (corresponding to between -30 and +30 days), with an equal probability of all 7 possible shifts occurring (including a shift by 0 timesteps). Random shifts are applied to all samples in the training and validation sets and differ for each sample every time it’s seen by the model. No shifts are applied to the samples in the testing sets. 

The primary metric for performance evaluation is the F\textsubscript{1} score on the test datasets of regions withheld from training. Accordingly, performance is assessed in a manner that prioritizes classifier robustness – i.e. performance in regions unseen during training – and not in a manner that could be inflated by close similarity of samples within a region. For reference, the F\textsubscript{1} score balances prediction precision and recall, and is calculated per Eq. \ref{f1}.

\begin{equation}
F_1 = \frac{TP}{TP+\frac{1}{2}(FP+FN)}
\label{f1}
\end{equation}

The training strategy differs for the tree-based classifiers and for the neural network-based classifiers. As training the tree-based classifiers occurs across a single batch with no iteration across epochs, there is no need for separate validation and testing datasets: The training and validation datasets of all included regions are therefore combined to create a single training dataset. After training on this combined dataset, performance is evaluated across the test datasets. 

In contrast, training neural network-based models takes place by batch across epochs, and a validation set is required to guide the training process. For a given training step, one batch from each region is concatenated, with the combined output shuffled before model intake. After the epoch is finished, performance is assessed on the validation set of each region included in training. If the minimum F\textsubscript{1} score among all regions’ validation sets has increased from its previous maximum, the model weights are saved; however, if the minimum F\textsubscript{1} score has not increased from its previous high point, the model weights are discarded. Minimum F\textsubscript{1} score across all validation regions is selected as the weight update criteria to ensure model robustness: Consistent performance across the entire area of interest is desired, not high performance in one set of regions and poor performance in another. Training concludes once the minimum validation region F\textsubscript{1} score has not improved for 10 training epochs, or after 30 epochs have been completed. After training, model weights are loaded from the epoch with the highest minimum validation region F\textsubscript{1} score; performance of this model on the test datasets of all regions is then reported. For all training runs, a binary cross-entropy loss, a learning rate of 1e-4, and an Adam optimizer \citep{Kingma2015} are specified. Inputs are standardized to a mean of 0 and standard deviation of 1 using statistics from the entire set of labeled samples. 

\section{Results}
\subsection{Model sensitivity}

\begin{figure}[ht]
\begin{center}
\includegraphics[width=13cm]{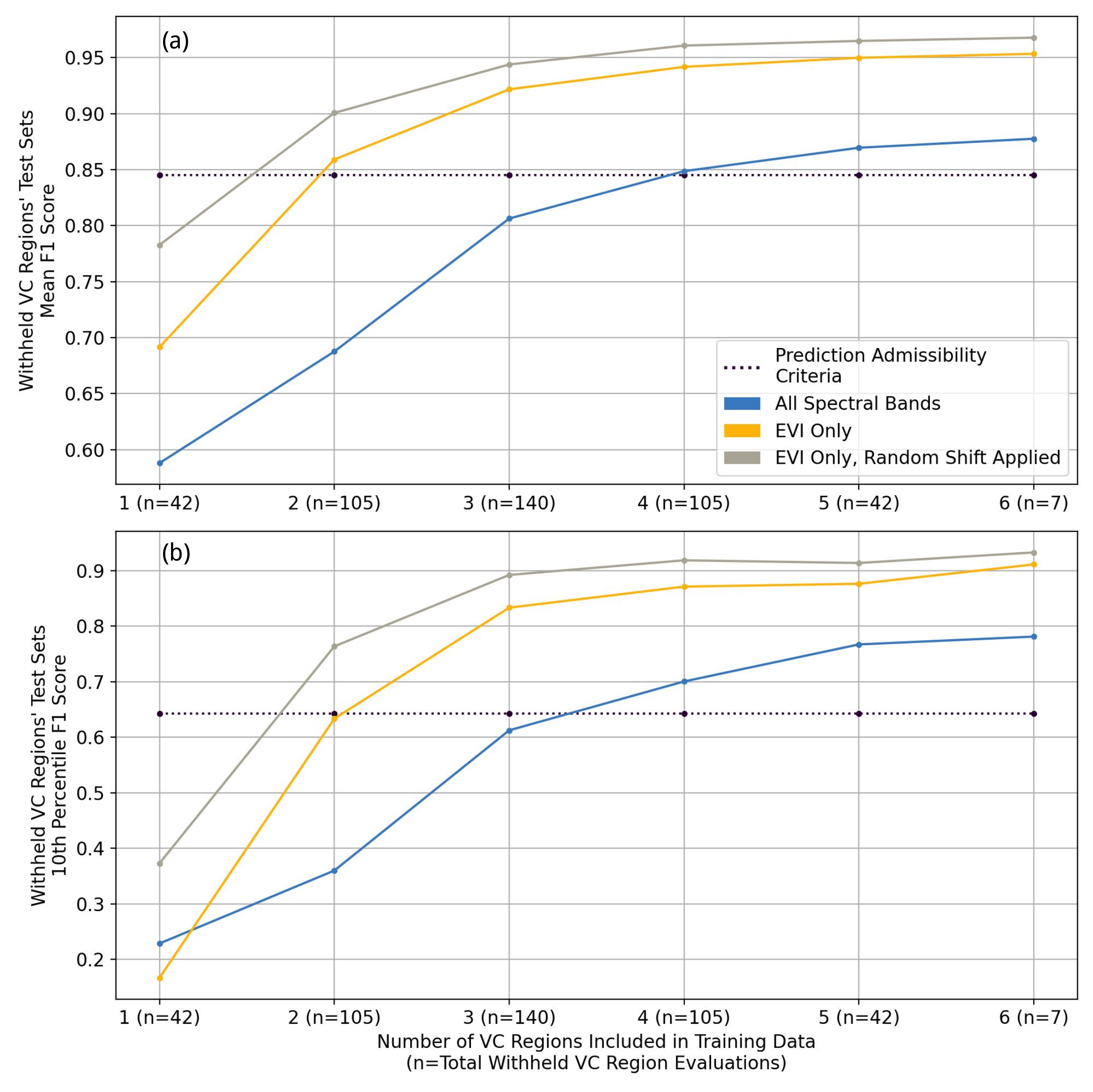}
\end{center}
\caption{Withheld region test dataset performance for different types of model input, organized along the \textit{x}-axis by the number of regions included during training. (a) presents mean F\textsubscript{1} score over the withheld regions; (b) presents the 10\textsuperscript{th} percentile F\textsubscript{1} score over the withheld regions. Results indicate that model inputs of randomly shifted enhanced vegetation index (EVI) timeseries yield the best classifier performance. F\textsubscript{1} scores from classification based on the prediction admissibility criteria are presented for reference.}
\label{model_input_performance}
\end{figure}

Figure \ref{model_input_performance} presents withheld VC region test dataset F\textsubscript{1} scores for three different types of model input – one that includes all spectral bands for all timesteps; one that includes only the EVI layer for all timesteps; and one that includes only the EVI layer for all timesteps with the random sample shift applied. Here, the performance of models trained on all combinations of VC regions is evaluated; these results are organized along the \textit{x}-axis by the number of VC regions included during training. Each \textit{x}-axis tick label also includes in parentheses the number of withheld VC region test dataset evaluations, \textit{n}, for all models trained on \textit{x} included VC regions\footnote{An example helps explain the calculation of \textit{n} values: Given $x=2$ VC regions included in training, there remain 5 VC regions unseen by the classifier. As there are ${7\choose2}=21$ ways to select 2 VC regions from the full set of 7, and each of these combinations leaves 5 withheld VC regions for performance evaluation, $n=105$ when $x=2$.}. Mean and 10\textsuperscript{th} percentile values of the \textit{n} performance evaluations are displayed for each \textit{x} between 1 and 6. All results are presented for the transformer model architecture; however, these findings are agnostic to the classifier architecture selected. 

Figure \ref{model_input_performance} demonstrates that models trained on samples containing only EVI timeseries outperform those that include all spectral bands at all timesteps, both on average (a) and in low performing regions  (b). The 10\textsuperscript{th} percentile of withheld regions’ F\textsubscript{1} scores is shown in order to understand the low-end of model performance without accounting for outliers. For reference, classifier performance based on the prediction admissibility criteria is also included. Figure \ref{model_input_performance} shows that explicitly feeding classification models information about samples’ vegetation content – i.e. feature engineering – allows for better performance compared to models that intake 10 Sentinel-2 spectral bands. Introducing a random temporal shift to the EVI timeseries further increases performance; by increasing the sample variance seen by the model, randomly shifting the input timeseries improves model transferability. Supplementary Figure S4 provides additional evidence of the benefits of this training strategy: A gradient class-activation map shows that a classifier trained on randomly shifted timeseries better identifies dry season vegetation as predictive of irrigation presence.

\begin{figure}[ht]
\begin{center}
\includegraphics[width=13cm]{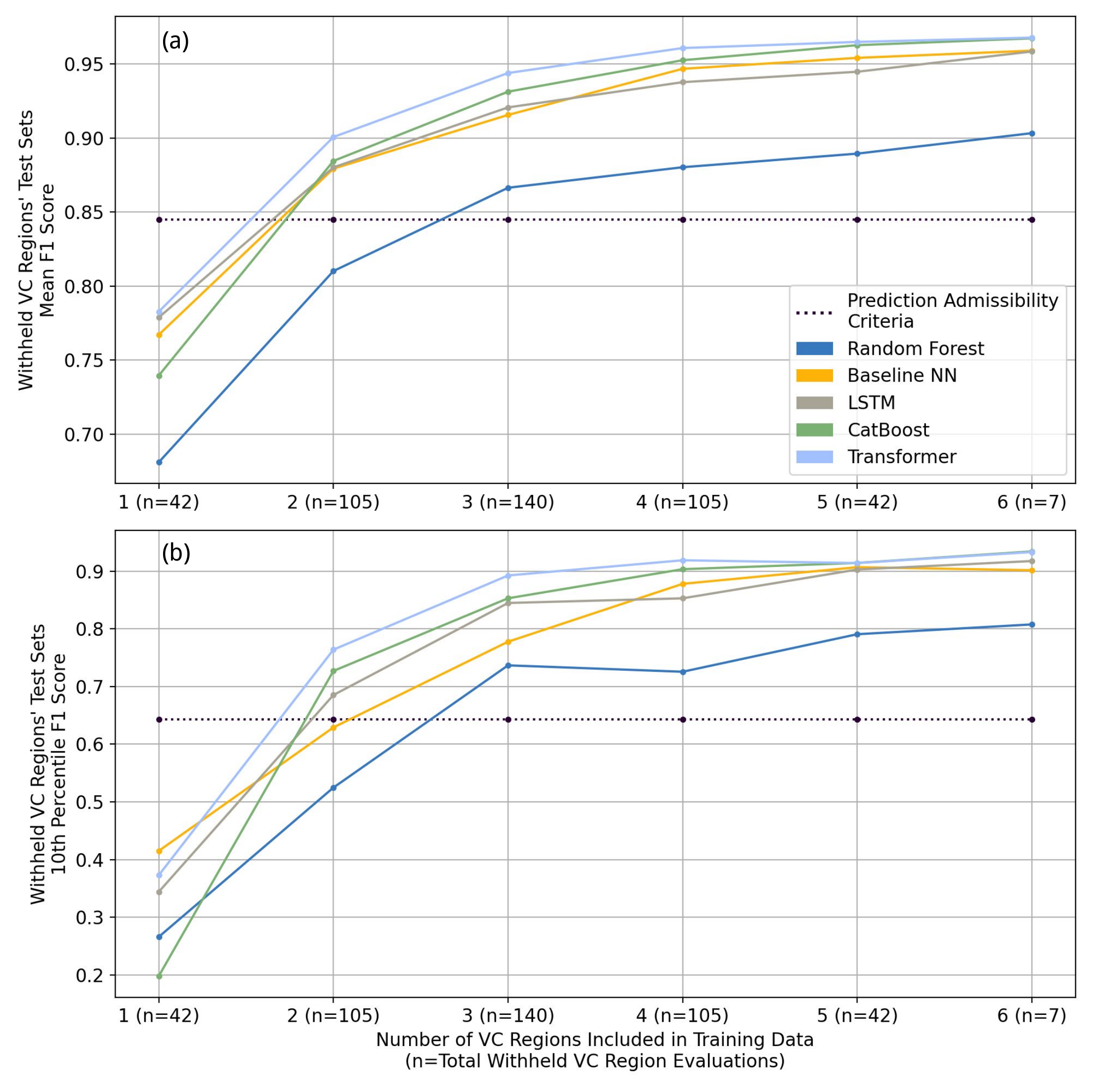}
\end{center}
\caption{Withheld region test dataset performance for different classifier models, organized along the \textit{x}-axis by the number of regions included during training. (a) presents mean F\textsubscript{1} score over the withheld regions; (b) presents the 10\textsuperscript{th} percentile F\textsubscript{1} score over the withheld regions. Results indicate that the transformer based classifier yields the best performance, followed closely by the CatBoost model. F\textsubscript{1} scores from classification based on the prediction admissibility criteria are presented for reference.}\label{model_architecture_performance}
\end{figure}

Taken together, randomly shifted EVI timeseries increase withheld region F\textsubscript{1} scores by an average of 0.22 when only 2 VC regions are included in the training data, compared to models that use all spectral bands. As performance begins to plateau with 4 or more VC regions included in the training data, this gap shrinks to an improvement of 0.10. Similar results can be seen in Figure \ref{model_input_performance}(b) for the low-end of performance: Extracting and randomly shifting EVI timeseries increase the 10\textsuperscript{th} percentile of withheld region F\textsubscript{1} scores by 0.40 when 2 VC regions are included in the training data, a difference that shrinks to approximately 0.14 with 5 or more VC regions in the training data. Two additional findings are gleaned from the results for the models trained on randomly shifted EVI timeseries (i.e. the grey curve). First, a classifier trained on data from 2 VC regions or more outperforms the pixel filtering baseline. Second, increasing the number of VC regions included in the training set improves withheld region prediction performance up until 4 VC regions before tapering off. 

Figure \ref{model_architecture_performance} displays (a) mean and (b) 10\textsuperscript{th} percentile F\textsubscript{1} score for all combinations of VC regions included in training for the 5 classification models tested, along with the reference classifier based on the prediction admissibility criteria. Figure \ref{model_architecture_performance} demonstrates that the transformer architecture is most robust for all combinations of VC training regions, followed closely by the CatBoost architecture for all training configurations with 2 or more VC regions. Moreover, for models with 5 or 6 VC regions included in training, mean and low-end F\textsubscript{1} scores for these two architectures are practically indistinguishable at 0.97 and 0.92, respectively. The Supplementary Materials contain further comparisons between Transformer and CatBoost performance (see Supplementary Table S6), showing that when each model is trained on all 7 VC regions’ training data, the two models demonstrate an average regional prediction alignment of 98.9\%. Moreover, an ablation study on training dataset size finds that reducing the proportion of polygons in the training set from 70\% to 15\% has minimal impact on prediction performance (See Supplementary Figure S5). Lastly, Figure \ref{model_architecture_performance} shows that the LSTM architecture does not noticeably improve performance compared to the baseline neural network, and that the trained Random Forest models yield the worst performance in withheld regions.

Next, prediction performance over the unseen ground-collected samples in Tana is assessed. As the transformer model demonstrates the most robust performance over withheld regions' samples, it is selected for prediction, achieving 96.7\% accuracy over irrigated samples (88,128/91,898) and 95.9\% accuracy over non-irrigated samples (33,954/35,121) for an F\textsubscript{1} score of 0.932. It is again worth noting that these high accuracies are achieved over the GC samples without the classification model seeing any labeled data from the Tana region during training.  

\subsection{Model inference}

For model inference, the transformer architecture is trained on the randomly shifted EVI timeseries of the labeled data from the 7 VC and one GC regions. The trained model is then deployed over Tigray and Amhara for the 2020 and 2021 dry seasons (using imagery collected between June 1, 2019 and June 1, 2020; and between June 1, 2020 and June 1, 2021, respectively). Two post-processing steps are then taken: 1) the prediction admissibility criteria are applied, and 2) contiguous groups of predicted irrigated pixels smaller than 0.1 Ha are removed in order to ignore isolated, outlier predictions. 

During inference, another step is taken to verify the accuracy of irrigation predictions. Here, five additional enumerators collect 1601 labeled polygons for the 2020 and 2021 dry seasons – 1082 non-irrigated polygons covering 3807 Ha and 519 irrigated polygons covering 582 Ha – across the extent of Amhara via the same labeling methodology used to collect the training, validation, and testing data. The locations of these independently labeled polygons are shown in Supplementary Figure S6. After cluster cleaning and applying the prediction admissibility criteria, these polygons yield 361,451 non-irrigated samples and 48,465 irrigated samples. An F\textsubscript{1} score of 0.917 is achieved over these samples – 98.3\% accuracy over non-irrigated samples and 95.5\% accuracy over irrigated samples, performance that remains in line with the reported test dataset metrics from Figure \ref{model_architecture_performance} and accuracies over the withheld Tana ground-collected labels.

Due to text constraints, Figures \ref{tigray_map} and \ref{amhara_map} present bitemporal irrigation maps at a resolution far coarser than their native 10m. The full resolution, georeferenced irrigation maps are available from the corresponding author upon request.

\subsubsection{Tigray}

\begin{figure}[ht]
\begin{center}
\includegraphics[width=\textwidth]{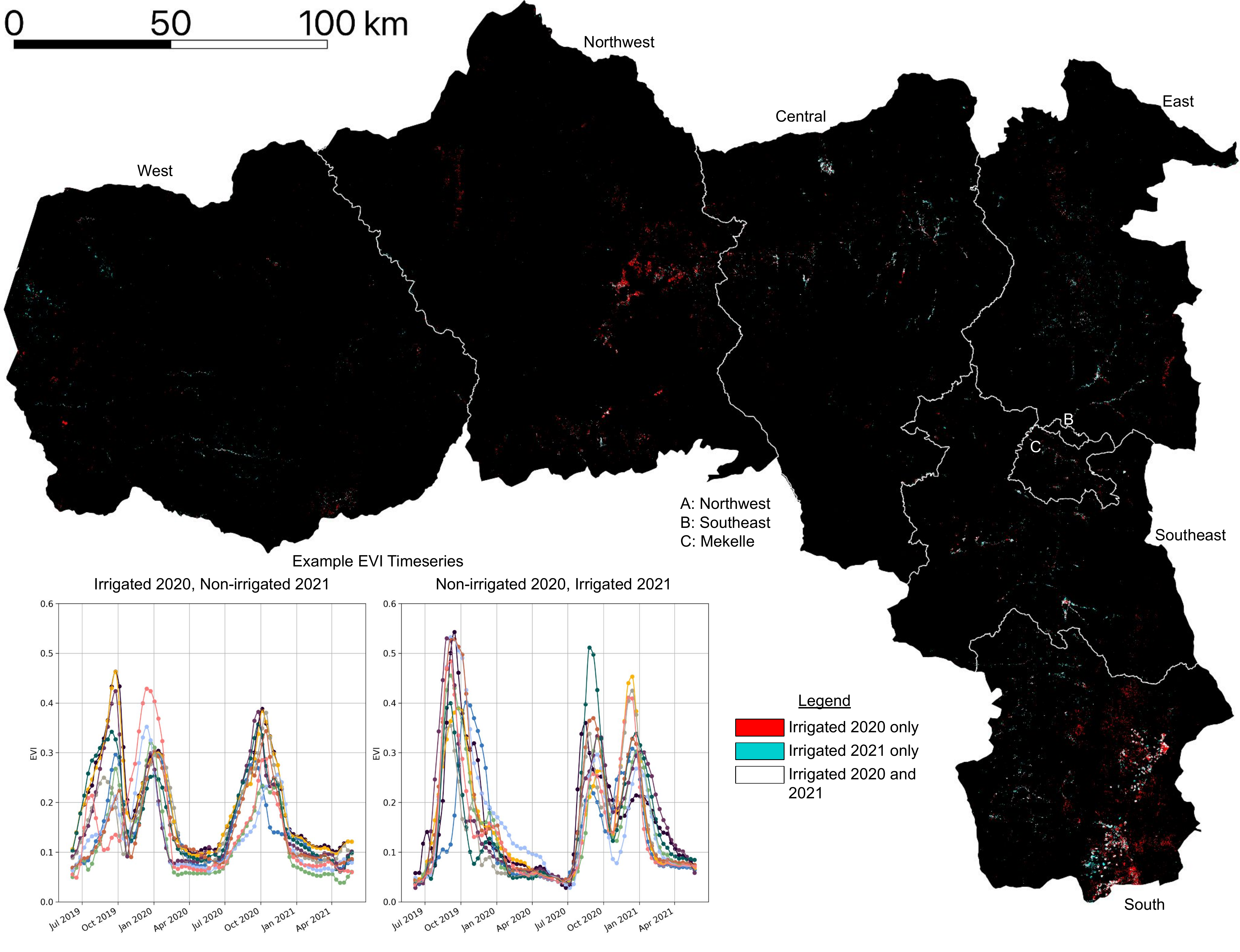}
\end{center}
\caption{Bitemporal irrigation map for Tigray. Figure inset contains example EVI timeseries predicted as irrigated in either 2020 or 2021. A predominance of red indicates that many parts of Tigray contain irrigation detected in 2020 but not in 2021.} 
\label{tigray_map}
\end{figure}

Figure \ref{tigray_map} presents predicted irrigated areas in Tigray for 2020 and 2021, with 2020 irrigation predictions in red and 2021 irrigation predictions in cyan. To better understand the nature of changing vegetation phenologies across this time period, the inset of Figure \ref{tigray_map} contains example timeseries that produced an irrigation prediction in one of 2020 or 2021. These example timeseries show that a second crop cycle with vegetation growth peaking in January is associated with a positive irrigation prediction; in contrast, the non-existence of this cycle is associated with non-irrigated prediction. Table \ref{tab:tigray_irrig_stats} displays the total predicted irrigated area for Tigray for 2020 and 2021, along with the total land area, organized by zone. Between 2020 and 2021, Table \ref{tab:tigray_irrig_stats} quantifies a 39.8\% decline in irrigated area in Tigray.

\begin{table}[h]
\scriptsize
 \caption{Predicted irrigated area statistics in Tigray for 2020 and 2021, organized by zone.}
  \centering
  \begin{tabular}{lrrrcc}
    \toprule
     Zone & \makecell[cr]{Irrigated Ha.,\\2020} & \makecell[cr]{Irrigated Ha.\\2021} & Total Ha. & \makecell[cc]{Percent Change,\\2020 to 2021} & \makecell[cc]{Percent Change as Fraction\\of Total Area, 2020 to 2021}\\ \midrule \midrule
     Central & 3710 &  3554 & 954,616 & -4.2\% & 0.0\% \\ \midrule
    Eastern &  3068 &  2863 & 635,670 & -6.7\% & 0.0\% \\ \midrule
    Mekelle & 556 & 397 & 52,313 & -28.5\% & -0.3\% \\ \midrule
    North Western & 7439 & 2062 & 1,246,715 & -72.3\% & -0.4\% \\    \midrule  
    South Eastern & 2658 & 2301 & 533,334 & -13.4\% & -0.1\% \\ \midrule
    Southern &  16,474 & 8064 & 506,151 & -51.1\% & -1.7\% \\ \midrule
    Western &  2278 & 2557 & 1,331,652 & 12.3\% & 0.0\% \\\midrule \midrule
    Total & 36,181 & 21,799 & 5,260,451 & -39.8\% & -0.3\% \\
    
    \bottomrule
  \end{tabular}
  \label{tab:tigray_irrig_stats}
\end{table}

\subsubsection{Amhara}

\begin{figure}[ht]
\begin{center}
\includegraphics[width=\textwidth]{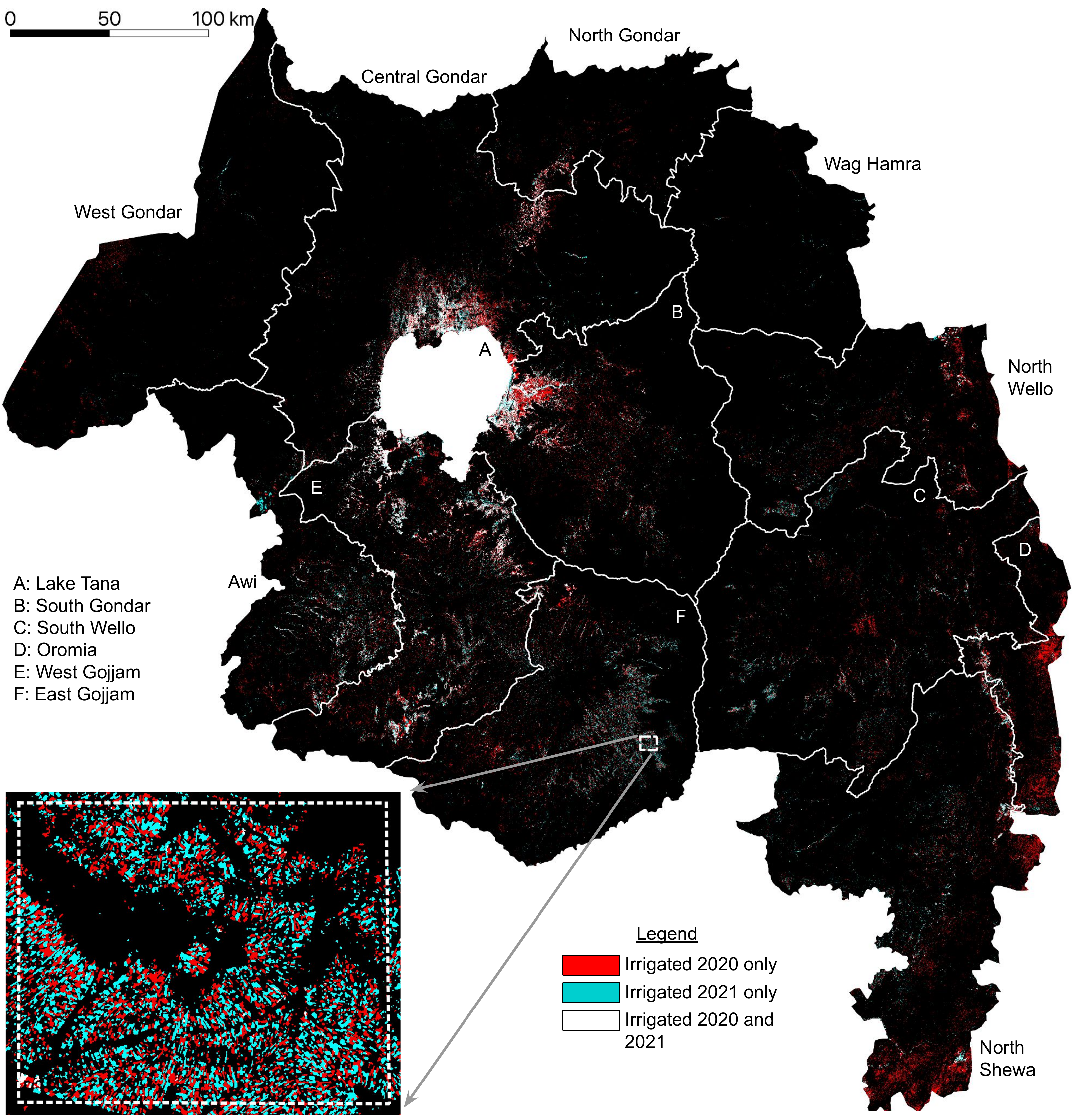}
\end{center}
\caption{Bitemporal irrigation map for Amhara. Figure inset contains example predictions around Choke Mountain displaying interannual irrigation patterns. A predominance of red indicates that many parts of Amhara contain irrigation detected in 2020 but not in 2021.}\label{amhara_map}
\end{figure}

Figure \ref{amhara_map} presents a bitemporal irrigation map for Amhara, also with 2020 irrigation predictions in red and 2021 irrigation predictions in cyan. This map contains large clusters of irrigated predictions around Lake Tana in the zones of Central Gondar, South Gondar, and West Gojjam, an intuitive finding given the availability of water from Lake Tana and the rivers that extend off it. Irrigation is also detected in the portions of Amhara’s easternmost zones that fall within the Main Ethiopian Rift (MER); as the valley formed by the MER extends north into Tigray, irrigation predictions in the North Wello, Oromia, and North Shewa zones align with irrigation predictions in the Southern zone of Tigray shown in Figure \ref{tigray_map}. Table \ref{tab:amhara_irrig_stats} displays the total predicted irrigated area for Amhara for 2020 and 2021, along with the total land area, organized by zone. From 2020 to 2021, Table \ref{tab:amhara_irrig_stats} quantifies a 41.6\% decline in irrigated area in Amhara.

The inset of Figure \ref{amhara_map} presents interannual irrigated cropping patterns for an area southwest of Choke Mountain. Interlocking red and cyan plots indicate the spatial rotation of irrigated crops from 2020 to 2021; no white plots are observed, which would signify dry season crop growth in both years.

\begin{table}[h]
\scriptsize
 \caption{Predicted irrigated area statistics in Amhara for 2020 and 2021, organized by zone.}
  \centering
  \begin{tabular}{lrrrcc}
    \toprule
     Zone & \makecell[cr]{Irrigated Ha.,\\2020} & \makecell[cr]{Irrigated Ha.\\2021} & Total Ha. & \makecell[cc]{Percent Change,\\2020 to 2021} & \makecell[cc]{Percent Change as Fraction\\of Total Area, 2020 to 2021}\\ \midrule \midrule
     
     Awi & 27,443 & 20,547 & 906,682 & -25.1\% & -0.8\%  \\ \midrule
Central Gondar & 73,450 & 50,954 & 2,095,018 & -30.6\% &-1.1\% \\ \midrule
East Gojjam & 44,975 & 33,888 & 1,405,689 & -24.7\% &-0.8\% \\ \midrule
North Gondar & 7381 & 3367 & 684,247 & -54.4\% &-0.6\% \\ \midrule
North Shewa (AM) & 62,933 & 21,362 & 1,622,197 & -66.1\% &-2.6\% \\ \midrule
North Wello & 21,367 & 8250 & 1,110,856 & -61.4\% &-1.2\% \\ \midrule
Oromia & 30,875 & 5285 & 380,773 & -82.9\% &-6.7\% \\ \midrule
South Gondar & 72,682 & 43,046 & 1,406,698 & -40.8\% &-2.1\% \\ \midrule
South Wello & 28,215 & 16,302 & 1,849,812 & -42.2\% &-0.6\% \\ \midrule
Wag Hamra & 447 & 698 & 890,004 & 56.4\% & 0.0\% \\ \midrule
West Gojjam & 97,206 & 71,052 & 1,348,157 & -26.9\% &-1.9\% \\ \midrule
West Gondar & 6180 & 1342 & 1,529,197 & -78.3\% &-0.3\% \\ \midrule \midrule
Total & 473,155 & 276,093 & 15,229,329 & -41.6\% &-1.3\% \\ \midrule
     
    \bottomrule
  \end{tabular}
  \label{tab:amhara_irrig_stats}
\end{table}

\section{Discussion}

This manuscript makes a set of contributions to the literature for learning from limited labels. First, it demonstrates a process of collecting training data to supplement ground-collected labels that improves on previous methods of sample collection -- such as using imagery from a single timestep or simple vegetation content heuristics -- as it verifies the existence or non-existence of full vegetation cycles during the dry season. Second, an evaluation of inputs, classifier architectures, and training strategies is presented for achieving irrigation classifier applicability to a larger area. Results indicate that enhanced vegetation (EVI) timeseries outperform a full set of spectral bands as inputs; that randomly shifting input timeseries prevents classifier models from overfitting to region-specific input features; and that a transformer-based neural network produces the highest prediction accuracies in unseen target regions. Due to the close similarity of performance metrics and alignment of predictions, the faster training, more easily interpretable CatBoost architecture is also shown as a suitable alternative for irrigation mapping efforts.

Prediction results indicate strong classifier performance over sample timeseries from regions not seen during training. On data from withheld target regions, transformer-based classifiers achieve mean F\textsubscript{1} scores above 0.95 when four or more regions' data are included during training; using labels from all 7 visual collection (VC) regions, the transformer-based classifier achieves an  F\textsubscript{1} score of 0.932 on the ground collection (GC) labels around Lake Tana. Over an independently collected set of more than 400,000 samples collected for performance assessment, the same classifier achieves 98.3\% accuracy over non-irrigated samples and 95.5\% accuracy over irrigated samples, demonstrating strong performance throughout the entire Ethiopian highlands.

Deploying a transformer-based classifier trained on samples from all 8 label collection regions yields insight into changing irrigation patterns. Results suggest that from 2020 to 2021, irrigation in Tigray and Amhara decreased by 40\%. In Tigray, this decline was most precipitous in the Northwest and Southern zones, which saw percent changes in irrigated area of -72.3\% and -51.1\%, respectively. The Western zone of Tigray was the only zone to see an increase in irrigated area from 2020 to 2021; even so, this increase amounted to 279 Ha in a zone with a total area of 1,331,652 Ha. Amhara is predicted to have had similar decreases in irrigated area: Apart from the Wag Hamra zone, which was predicted to have less than 0.08\% of its area irrigated in 2020 or 2021, all zones in Amhara experienced a change in irrigated area between -25.0\% and -82.3\%. The largest declines by area occurred in North Shewa (-41,572 Ha), South Gondar (-29,636 Ha), and West Gojjam (-26,154 Ha). Combined, results for Tigray and Amhara predict severe reductions in dry season crop growth from 2020 to 2021, findings that that align with recent reports of food insecurity following the eruption of civil conflict in Ethiopia in late 2020. 

Despite presented performance metrics indicating high levels of prediction accuracy, there are a few limitations to the proposed methodology that are important to mention. First, the study area is limited to the Ethiopian highlands, a highly agricultural, climatologically consistent area that is dominated by rainfed cropped phenologies. As the irrigation classifiers are only trained to separate dry season crop cycles from rainfed vegetation cycles -- associating identified dry-season cropping with irrigation presence -- they will perform poorly in settings with different rainfall and phenological patterns. Relatedly, the trained irrigation classifiers do not identify irrigation used to supplement rainy season precipitation, irrigation of perennial tree crops, evergreen vegetation in riparian areas, or irrigation that supports continuous cropping, as the phenological signatures of these types of vegetation are difficult to distinguish from evergreen, non-cropped signatures. This discrimination task is left for future work. Lastly, classifiers are trained only on cropped phenologies, which constitute a portion of the vegetation signatures that exist in the area of interest. To manage the other phenologies present at model inference, prediction admissibility criteria are implemented. Nevertheless, these criteria are imperfect: There are surely irrigated pixels which have been mistakenly assigned a non-irrigated class label, along with non-cropped pixels which have evaded the admissibility criteria.

While the presented methodology is applied only for the task of irrigation identification in the Ethiopian highlands, the strategy of regional phenological characterization to provide context for geographically informed selection of training samples and model applicability can be transferred more broadly to a range of land process mapping objectives. The suitability of this approach in the field of machine learning with limited labels is supported by results comparing classifier architectures and hyperparameter choice to assess the question of result uniqueness that overshadows all land cover classifications. As discussed by \citet{Small2021}, what is presented as \textit{the} map is often just \textit{a} map -- one of many different products that can be obtained from the same set of inputs with different classifiers and hyperparameter settings. By assessing multiple classifier architectures and quantifying prediction sensitivity, this approach demonstrates consistency in results and indicates the uncertainty that can be expected of the resulting irrigation maps; as such, it provides a process for building robust classifiers in settings with scarce labeled data.

\section*{Conflict of Interest Statement}

The authors declare that the research was conducted in the absence of any commercial or financial relationships that could be construed as a potential conflict of interest.

\section*{Author Contributions}

TC and VM conceived of the study, which was led by VM. TC developed and implemented the methodology, analyzed the results, and produced the data visualizations. CS introduced the concept of multiscale phenological context and devised the spatiotemporal mixture model. VM consulted in all steps of these processes. TC is the primary author of the manuscript, which was prepared with editorial assistance from CS and VM.

\section*{Funding}
Partial support for this effort was provided by the National Science Foundation (INFEWS Award Number 1639214), Columbia World Projects, Rockefeller Foundation (eGuide Grant 2018POW004), OPML UK (DFID) and Technoserve (BMGF). 

\section*{Acknowledgments}

The authors are grateful to Jack Bott, Yinbo Hu, Hasan Siddiqui, and Yuezi Wu for their assistance in labeling. The authors would like to thank Gunther Bensch (RWI), Andrej Kveder (OPML), Abiy Tamerat (EthioResource Group), Yifru Tadesse (ATA Ethiopia), and Esther Kim (Technoserve) for their assistance with field data collection efforts; Rose Rustowicz for guidance in using of Descartes Labs platform; and colleagues Jay Taneja (UMass Amherst), Markus Walsh (AfSIS), and Edwin Adkins (Columbia) for their continued stimulating discussions and guidance.  

\section*{Data Availability Statement}
Labeled data and predicted irrigation maps are available from the corresponding author upon request.

\bibliographystyle{abbrvnat}
\bibliography{references}  

\begin{thebibliography}{36}
\providecommand{\natexlab}[1]{#1}
\providecommand{\url}[1]{\texttt{#1}}
\expandafter\ifx\csname urlstyle\endcsname\relax
  \providecommand{\doi}[1]{doi: #1}\else
  \providecommand{\doi}{doi: \begingroup \urlstyle{rm}\Url}\fi

\bibitem[Abbasi et~al.(2015)Abbasi, Arefi, Bigdeli, and Roessner]{Abbasi2015}
B.~Abbasi, H.~Arefi, B.~Bigdeli, and S.~Roessner.
\newblock {Automatic generation of training data for hyperspectral image
  classification using support vector machine}.
\newblock \emph{International Archives of the Photogrammetry, Remote Sensing
  and Spatial Information Sciences - ISPRS Archives}, 40\penalty0
  (7W3):\penalty0 575--580, 2015.
\newblock \doi{10.5194/isprsarchives-XL-7-W3-575-2015}.

\bibitem[Banerjee et~al.(2015)Banerjee, Bovolo, Bhattacharya, Bruzzone,
  Chaudhuri, and Mohan]{Banerjee2015}
B.~Banerjee, F.~Bovolo, A.~Bhattacharya, L.~Bruzzone, S.~Chaudhuri, and B.~K.
  Mohan.
\newblock {A new self-training-based unsupervised satellite image
  classification technique using cluster ensemble strategy}.
\newblock \emph{IEEE Geoscience and Remote Sensing Letters}, 12\penalty0
  (4):\penalty0 741--745, 2015.
\newblock ISSN 1545598X.
\newblock \doi{10.1109/LGRS.2014.2360833}.

\bibitem[Bazzi et~al.(2020)Bazzi, Baghdadi, Fayad, Zribi, Belhouchette, and
  Demarez]{Bazzi2020}
H.~Bazzi, N.~Baghdadi, I.~Fayad, M.~Zribi, H.~Belhouchette, and V.~Demarez.
\newblock {Near real-time irrigation detection at plot scale using sentinel-1
  data}.
\newblock \emph{Remote Sensing}, 12\penalty0 (9), 2020.
\newblock ISSN 20724292.
\newblock \doi{10.3390/RS12091456}.

\bibitem[Bazzi et~al.(2021)Bazzi, Baghdadi, Amin, Fayad, Zribi, Demarez, and
  Belhouchette]{Bazzi2021}
H.~Bazzi, N.~Baghdadi, G.~Amin, I.~Fayad, M.~Zribi, V.~Demarez, and
  H.~Belhouchette.
\newblock {An operational framework for mapping irrigated areas at plot scale
  using sentinel‐1 and sentinel‐2 data}.
\newblock \emph{Remote Sensing}, 13\penalty0 (13):\penalty0 1--28, 2021.
\newblock ISSN 20724292.
\newblock \doi{10.3390/rs13132584}.

\bibitem[Breiman(2001)]{Breiman2001}
L.~Breiman.
\newblock {Random Forests}.
\newblock \emph{Machine Learning}, pages 1--28, 2001.
\newblock \doi{10.1201/9780429469275-8}.

\bibitem[Chen et~al.(2018)Chen, Lu, Luo, Pokhrel, Deb, Huang, and
  Ran]{Chen2018}
Y.~Chen, D.~Lu, L.~Luo, Y.~Pokhrel, K.~Deb, J.~Huang, and Y.~Ran.
\newblock {Detecting irrigation extent, frequency, and timing in a
  heterogeneous arid agricultural region using MODIS time series, Landsat
  imagery, and ancillary data}.
\newblock \emph{Remote Sensing of Environment}, 204\penalty0 (March
  2017):\penalty0 197--211, 2018.
\newblock ISSN 00344257.
\newblock \doi{10.1016/j.rse.2017.10.030}.
\newblock URL \url{https://doi.org/10.1016/j.rse.2017.10.030}.

\bibitem[{Conlon} et~al.(2020){Conlon}, {Wu}, {Small}, {Siddiqui}, {Adkins},
  and {Modi}]{conlon2020}
T.~{Conlon}, Y.~{Wu}, C.~{Small}, H.~{Siddiqui}, E.~{Adkins}, and V.~{Modi}.
\newblock {A Novel Method of Irrigation Detection and Estimation of the Effects
  of Productive Electricity Demands on Energy System Planning}.
\newblock In \emph{AGU Fall Meeting Abstracts}, volume 2020, pages GC034--08,
  Dec. 2020.

\bibitem[de~Lima and Marfurt(2020)]{DeLima2020}
R.~P. de~Lima and K.~Marfurt.
\newblock {Convolutional neural network for remote-sensing scene
  classification: Transfer learning analysis}.
\newblock \emph{Remote Sensing}, 12\penalty0 (1), 2020.
\newblock ISSN 20724292.
\newblock \doi{10.3390/rs12010086}.

\bibitem[Deng et~al.(2009)Deng, {Wei Dong}, Socher, {Li-Jia Li}, {Kai Li}, and
  {Li Fei-Fei}]{JiaDeng2009}
J.~Deng, {Wei Dong}, R.~Socher, {Li-Jia Li}, {Kai Li}, and {Li Fei-Fei}.
\newblock {ImageNet: A large-scale hierarchical image database}.
\newblock pages 248--255, 2009.
\newblock \doi{10.1109/cvprw.2009.5206848}.

\bibitem[Dorogush et~al.(2017)Dorogush, Gulin, Gusev, Kazeev, Prokhorenkova,
  and Vorobev]{prokhorenkova2019catboost}
A.~V. Dorogush, A.~Gulin, G.~Gusev, N.~Kazeev, L.~O. Prokhorenkova, and
  A.~Vorobev.
\newblock Fighting biases with dynamic boosting.
\newblock \emph{CoRR}, abs/1706.09516, 2017.
\newblock URL \url{http://arxiv.org/abs/1706.09516}.

\bibitem[Gebregziabher et~al.(2009)Gebregziabher, Namara, and
  Holden]{Gebregziabher2009}
G.~Gebregziabher, R.~E. Namara, and S.~Holden.
\newblock {Poverty reduction with irrigation investment: An empirical case
  study from Tigray, Ethiopia}.
\newblock \emph{Agricultural Water Management}, 96\penalty0 (12):\penalty0
  1837--1843, 2009.
\newblock ISSN 03783774.
\newblock \doi{10.1016/j.agwat.2009.08.004}.

\bibitem[Huete et~al.(1999)Huete, Justice, and Van~Leeuwen]{huete1999modis}
A.~Huete, C.~Justice, and W.~Van~Leeuwen.
\newblock {MODIS Vegetation Index (MOD13}) algorithm theoretical basis
  document.
\newblock 3\penalty0 (213):\penalty0 295--309, 1999.

\bibitem[King and Zeng(2001)]{King2001}
G.~King and L.~Zeng.
\newblock {Logistic regression in rare events data}.
\newblock \emph{Political Analysis}, 9\penalty0 (2):\penalty0 137--163, 2001.
\newblock ISSN 15487660.
\newblock \doi{10.18637/jss.v008.i02}.

\bibitem[Kingma and Ba(2015)]{Kingma2015}
D.~P. Kingma and J.~L. Ba.
\newblock {Adam: A method for stochastic optimization}.
\newblock \emph{3rd International Conference on Learning Representations, ICLR
  2015 - Conference Track Proceedings}, pages 1--15, 2015.

\bibitem[Lecun et~al.(2015)Lecun, Bengio, and Hinton]{Lecun2015}
Y.~Lecun, Y.~Bengio, and G.~Hinton.
\newblock {Deep learning}.
\newblock \emph{Nature}, 521\penalty0 (7553):\penalty0 436--444, 2015.
\newblock ISSN 14764687.
\newblock \doi{10.1038/nature14539}.

\bibitem[Li et~al.(2018)Li, Zhang, Xue, Jiang, and Shen]{Li2018}
Y.~Li, H.~Zhang, X.~Xue, Y.~Jiang, and Q.~Shen.
\newblock {Deep learning for remote sensing image classification: A survey}.
\newblock \emph{Wiley Interdisciplinary Reviews: Data Mining and Knowledge
  Discovery}, 8\penalty0 (6):\penalty0 1--17, 2018.
\newblock ISSN 19424795.
\newblock \doi{10.1002/widm.1264}.

\bibitem[Naik and Kumar(2021)]{Naik2021}
P.~Naik and A.~Kumar.
\newblock \emph{{A Stochastic Approach for Automatic Collection of Precise
  Training Data for a Soft Machine Learning Algorithm Using Remote Sensing
  Images}}.
\newblock Springer Singapore, 2021.
\newblock \doi{10.1007/978-981-16-2712-5_24}.

\bibitem[Ozdogan et~al.(2010)Ozdogan, Yang, Allez, and Cervantes]{Ozdogan2010}
M.~Ozdogan, Y.~Yang, G.~Allez, and C.~Cervantes.
\newblock {Remote sensing of irrigated agriculture: Opportunities and
  challenges}.
\newblock \emph{Remote Sensing}, 2\penalty0 (9):\penalty0 2274--2304, 2010.
\newblock ISSN 20724292.
\newblock \doi{10.3390/rs2092274}.

\bibitem[Pervez et~al.(2014)Pervez, Budde, and Rowland]{ShahriarPervez2014}
M.~S. Pervez, M.~Budde, and J.~Rowland.
\newblock {Mapping irrigated areas in Afghanistan over the past decade using
  MODIS NDVI}.
\newblock \emph{Remote Sensing of Environment}, 149:\penalty0 155--165, jun
  2014.
\newblock ISSN 0034-4257.
\newblock \doi{10.1016/J.RSE.2014.04.008}.

\bibitem[Phiri et~al.(2020)Phiri, Simwanda, Salekin, Nyirenda, Murayama, and
  Ranagalage]{Phiri2019}
D.~Phiri, M.~Simwanda, S.~Salekin, V.~R. Nyirenda, Y.~Murayama, and
  M.~Ranagalage.
\newblock Sentinel-2 data for land cover/use mapping: A review.
\newblock \emph{Remote Sensing}, 12\penalty0 (14), 2020.
\newblock ISSN 2072-4292.
\newblock \doi{10.3390/rs12142291}.

\bibitem[Ramezan et~al.(2021)Ramezan, Warner, Maxwell, and Price]{Ramezan2021}
C.~A. Ramezan, T.~A. Warner, A.~E. Maxwell, and B.~S. Price.
\newblock {Effects of training set size on supervised machine-learning
  land-cover classification of large-area high-resolution remotely sensed
  data}.
\newblock \emph{Remote Sensing}, 13\penalty0 (3):\penalty0 1--27, 2021.
\newblock \doi{10.3390/rs13030368}.

\bibitem[Saha et~al.(2019)Saha, Solano-Correa, Bovolo, and Bruzzone]{Saha2019}
S.~Saha, Y.~T. Solano-Correa, F.~Bovolo, and L.~Bruzzone.
\newblock {Unsupervised deep learning based change detection in Sentinel-2
  images}.
\newblock \emph{2019 10th International Workshop on the Analysis of
  Multitemporal Remote Sensing Images, MultiTemp 2019}, pages 0--3, 2019.
\newblock \doi{10.1109/Multi-Temp.2019.8866899}.

\bibitem[Sivaraj et~al.(2022)Sivaraj, Kumar, Koti, and Naik]{Sivaraj2022}
P.~Sivaraj, A.~Kumar, S.~R. Koti, and P.~Naik.
\newblock {Effects of Training Parameter Concept and Sample Size in
  Possibilistic c-Means Classifier for Pigeon Pea Specific Crop Mapping}.
\newblock \emph{Geomatics}, 2\penalty0 (1):\penalty0 107--124, 2022.
\newblock \doi{10.3390/geomatics2010007}.

\bibitem[Small(2012)]{Small2012}
C.~Small.
\newblock {Spatiotemporal dimensionality and Time-Space characterization of
  multitemporal imagery}.
\newblock \emph{Remote Sensing of Environment}, 124:\penalty0 793--809, 2012.
\newblock ISSN 00344257.
\newblock \doi{10.1016/j.rse.2012.05.031}.

\bibitem[Small(2021)]{Small2021}
C.~Small.
\newblock {Grand Challenges in Remote Sensing Image Analysis and
  Classification}.
\newblock \emph{Frontiers in Remote Sensing}, 1\penalty0 (April):\penalty0
  1--4, 2021.
\newblock \doi{10.3389/frsen.2020.605220}.

\bibitem[Stivaktakis et~al.(2019)Stivaktakis, Tsagkatakis, and
  Tsakalides]{Stivaktakis2019}
R.~Stivaktakis, G.~Tsagkatakis, and P.~Tsakalides.
\newblock {Deep Learning for Multilabel Land Cover Scene Categorization Using
  Data Augmentation}.
\newblock \emph{IEEE Geoscience and Remote Sensing Letters}, 16\penalty0
  (7):\penalty0 1031--1035, 2019.
\newblock ISSN 15580571.
\newblock \doi{10.1109/LGRS.2019.2893306}.

\bibitem[Stromann et~al.(2020)Stromann, Nascetti, Yousif, and
  Ban]{Stromann2020}
O.~Stromann, A.~Nascetti, O.~Yousif, and Y.~Ban.
\newblock {Dimensionality Reduction and Feature Selection for Object-Based Land
  Cover Classification based on Sentinel-1 and Sentinel-2 Time Series Using
  Google Earth Engine}.
\newblock \emph{Remote Sensing}, 12\penalty0 (1), 2020.
\newblock \doi{10.3390/RS12010076}.

\bibitem[Tao et~al.(2020)Tao, Qi, Lu, Wang, and Li]{Tao2020}
C.~Tao, J.~Qi, W.~Lu, H.~Wang, and H.~Li.
\newblock Remote sensing image scene classification with self-supervised
  paradigm under limited labeled samples.
\newblock \emph{CoRR}, abs/2010.00882, 2020.
\newblock URL \url{https://arxiv.org/abs/2010.00882}.

\bibitem[Vogels et~al.(2019{\natexlab{a}})Vogels, de~Jong, Sterk, and
  Addink]{Vogels2019}
M.~F. Vogels, S.~M. de~Jong, G.~Sterk, and E.~A. Addink.
\newblock {Mapping irrigated agriculture in complex landscapes using SPOT6
  imagery and object-based image analysis – A case study in the Central Rift
  Valley, Ethiopia}.
\newblock \emph{International Journal of Applied Earth Observation and
  Geoinformation}, 75\penalty0 (May 2018):\penalty0 118--129,
  2019{\natexlab{a}}.
\newblock ISSN 1872826X.
\newblock \doi{10.1016/j.jag.2018.07.019}.

\bibitem[Vogels et~al.(2019{\natexlab{b}})Vogels, de~Jong, Sterk, Douma, and
  Addink]{Vogels2019a}
M.~F. Vogels, S.~M. de~Jong, G.~Sterk, H.~Douma, and E.~A. Addink.
\newblock {Spatio-temporal patterns of smallholder irrigated agriculture in the
  horn of Africa using GEOBIA and Sentinel-2 imagery}.
\newblock \emph{Remote Sensing}, 11\penalty0 (2), 2019{\natexlab{b}}.
\newblock ISSN 20724292.
\newblock \doi{10.3390/rs11020143}.

\bibitem[Wakjira et~al.(2021)Wakjira, Peleg, Anghileri, Molnar, Alamirew, Six,
  and Molnar]{Wakjira2021}
M.~T. Wakjira, N.~Peleg, D.~Anghileri, D.~Molnar, T.~Alamirew, J.~Six, and
  P.~Molnar.
\newblock {Rainfall seasonality and timing: implications for cereal crop
  production in Ethiopia}.
\newblock \emph{Agricultural and Forest Meteorology}, 310:\penalty0 108633,
  2021.
\newblock ISSN 01681923.
\newblock \doi{10.1016/j.agrformet.2021.108633}.

\bibitem[Wang et~al.(2017)Wang, Yan, and Oates]{Wang2017}
Z.~Wang, W.~Yan, and T.~Oates.
\newblock {Time series classification from scratch with deep neural networks: A
  strong baseline}.
\newblock \emph{Proceedings of the International Joint Conference on Neural
  Networks}, 2017-May:\penalty0 1578--1585, 2017.
\newblock \doi{10.1109/IJCNN.2017.7966039}.

\bibitem[Wiggins et~al.(2021)Wiggins, Glover, and Dorgan]{Wiggins2021}
S.~Wiggins, D.~Glover, and A.~Dorgan.
\newblock {Agricultural innovation for smallholders in sub-Saharan Africa}.
\newblock Technical Report July, 2021.
\newblock URL
  \url{https://webarchive.nationalarchives.gov.uk/ukgwa/20211030121337/https://degrp.odi.org/publication/agricultural-innovation-for-smallholders-in-sub-saharan-africa/}.

\bibitem[Wu and Chin(2016)]{Wu2016}
Y.~Wu and L.~S. Chin.
\newblock {A simplified training data collection method for sequential remote
  sensing image classification}.
\newblock \emph{4th International Workshop on Earth Observation and Remote
  Sensing Applications, EORSA 2016 - Proceedings}, pages 329--332, 2016.
\newblock \doi{10.1109/EORSA.2016.7552823}.

\bibitem[Yu et~al.(2017)Yu, Wu, Luo, and Ren]{Yu2017}
X.~Yu, X.~Wu, C.~Luo, and P.~Ren.
\newblock {Deep learning in remote sensing scene classification: a data
  augmentation enhanced convolutional neural network framework}.
\newblock \emph{GIScience and Remote Sensing}, 54\penalty0 (5):\penalty0
  741--758, 2017.
\newblock ISSN 15481603.
\newblock \doi{10.1080/15481603.2017.1323377}.

\bibitem[Zhong et~al.(2019)Zhong, Hu, Zhou, and Tao]{Zhong2019}
L.~Zhong, L.~Hu, H.~Zhou, and X.~Tao.
\newblock {Deep learning based winter wheat mapping using statistical data as
  ground references in Kansas and northern Texas, US}.
\newblock \emph{Remote Sensing of Environment}, 233:\penalty0 111411, nov 2019.
\newblock ISSN 0034-4257.
\newblock \doi{10.1016/J.RSE.2019.111411}.

\end{thebibliography}


\begin{thebibliography}{4}
\providecommand{\natexlab}[1]{#1}
\providecommand{\url}[1]{\texttt{#1}}
\expandafter\ifx\csname urlstyle\endcsname\relax
  \providecommand{\doi}[1]{doi: #1}\else
  \providecommand{\doi}{doi: \begingroup \urlstyle{rm}\Url}\fi

\bibitem[Hagen et~al.(2020)Hagen, Strube, Haide, Kahn, Jackson, and
  Hainje]{Hagen2020}
A.~Hagen, J.~Strube, I.~Haide, J.~Kahn, S.~Jackson, and C.~Hainje.
\newblock {A Proposed High Dimensional Kolmogorov-Smirnov Distance}.
\newblock \emph{Machine Learning and the Physical Sciences: Workshop at the
  34th Conference on Neural Information Processing Systems (NeurIPS)},
  \penalty0 (NeurIPS):\penalty0 1--6, 2020.
\newblock URL \url{https://ml4physicalsciences.github.io/2020/}.

\bibitem[Main-Knorn et~al.(2017)Main-Knorn, Pflug, Louis, Debaecker,
  M{\"{u}}ller-Wilm, and Gascon]{sen2cor}
M.~Main-Knorn, B.~Pflug, J.~Louis, V.~Debaecker, U.~M{\"{u}}ller-Wilm, and
  F.~Gascon.
\newblock {Sen2Cor for Sentinel-2}.
\newblock In L.~Bruzzone, editor, \emph{Image and Signal Processing for Remote
  Sensing XXIII}, volume 10427, pages 37--48. International Society for Optics
  and Photonics, SPIE, 2017.
\newblock \doi{10.1117/12.2278218}.
\newblock URL \url{https://doi.org/10.1117/12.2278218}.

\bibitem[Selvaraju et~al.(2020)Selvaraju, Cogswell, Das, Vedantam, Parikh, and
  Batra]{Selvaraju2020}
R.~R. Selvaraju, M.~Cogswell, A.~Das, R.~Vedantam, D.~Parikh, and D.~Batra.
\newblock {Grad-CAM: Visual Explanations from Deep Networks via Gradient-Based
  Localization}.
\newblock \emph{International Journal of Computer Vision}, 128\penalty0
  (2):\penalty0 336--359, 2020.
\newblock ISSN 15731405.
\newblock \doi{10.1007/s11263-019-01228-7}.

\bibitem[Small(2012)]{Small2012}
C.~Small.
\newblock {Spatiotemporal dimensionality and Time-Space characterization of
  multitemporal imagery}.
\newblock \emph{Remote Sensing of Environment}, 124:\penalty0 793--809, 2012.
\newblock ISSN 00344257.
\newblock \doi{10.1016/j.rse.2012.05.031}.

\end{thebibliography}

\end{document}


\maketitle

\section{Supplementary Background}

\subsection{Phenology mapping by temporal endmember unmixing}

\begin{figure}[ht]
\begin{center}
\includegraphics[width=15cm]{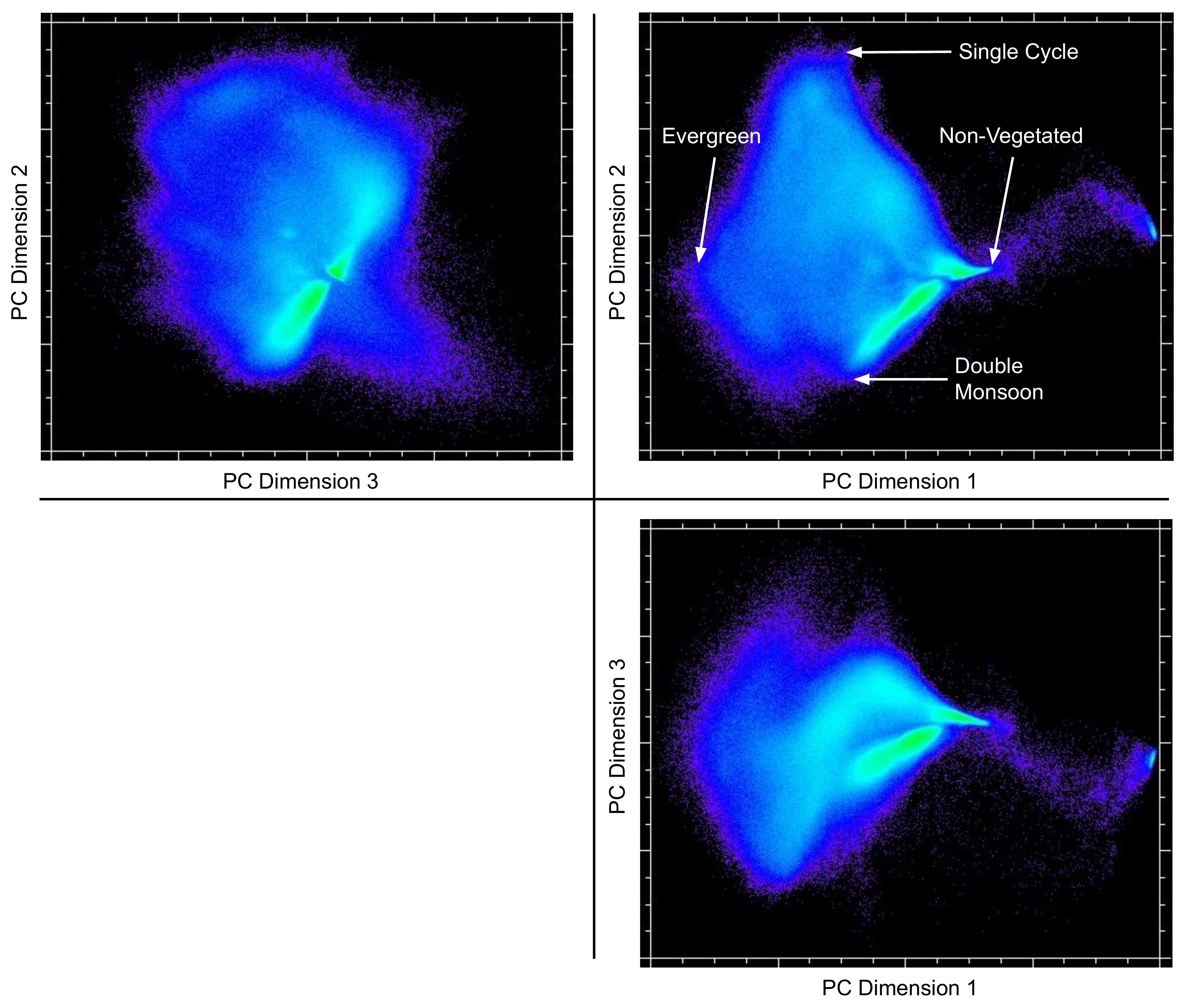}
\end{center}
\caption{Views of the first three principal component (PC) dimensions of the transformed 10-year Ethiopia MODIS enhanced vegetation index (EVI) imagery cube. Locations of temporal endmembers used to construct the phenology map in Figure 1 via unconstrained linear unmixing are presented in the PC Dimension 1 vs. PC Dimension 2 plot.}
\label{tem_scatter_plot}
\end{figure}

Figure 1 in the main text presents a vegetation phenology map for Ethiopia created by applying a temporal mixture model to an image cube containing 16-day 250m MODIS enhanced vegetation index (EVI) imagery between June 1, 2011 and June 1, 2021. Figure \ref{tem_scatter_plot} presents the locations of temporal endmember (tEM) extraction from the image cube transformed into principal component (PC) space. The four extracted tEMs are then used to create the phenology map via unconstrained least-squares linear unmixing per the methodology introduced in \citet{Small2012}. 

\begin{figure}[htp]
\begin{center}
\includegraphics[width=\textwidth]{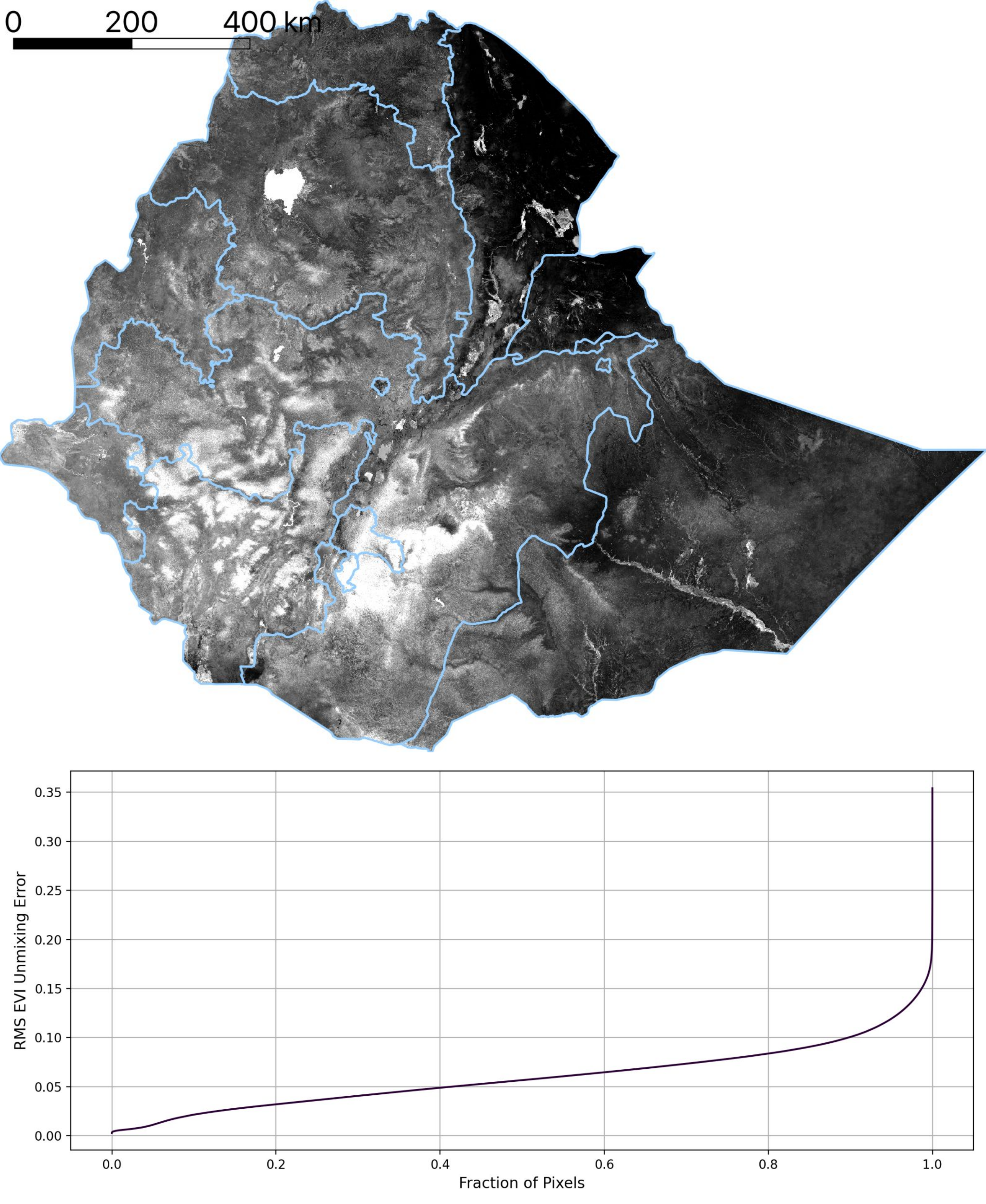}
\end{center}
\caption{Root mean square (RMS) error of the temporal mixture model inversion for Figure 1. A 2\% linear stretch is applied to the error map, with administrative boundaries outlined in light blue. The displayed cumulative density function shows the distribution of RMS errors for an increasing fraction of pixels in Figure 1.}
\label{tem_unmixing_error}
\end{figure}

Figure \ref{tem_unmixing_error} presents the temporal mixture model inversion error and the cumulative error statistics for Figure 1. Interpreting Figure 1 and Figure \ref{tem_unmixing_error} together reveals that the locations of highest error occur over evergreen vegetation, primarily in the southeast of Ethiopia. As the unmixing error remains low over Tigray and Amhara, the authors stipulate that Figure 1 contains an accurate assessment of vegetation cycles in the area of interest.

\section{Supplementary Methods}

\subsection{Sentinel-2 imagery collection}

Using the Descartes Labs platform, imagery mosaics are generating by collecting all Sentinel-2 imagery available within a 10-day timestep that come from 100km-by-100km granules with less than 10\% aggregate cloud cover. These images are then sorted by cloud cover and masked using the cloud masks provided by the European Space Agency (ESA) Sen2Cor algorithm \citep{sen2cor}.  Given the 5-day revisit period of Sentinel-2 near the equator, a 10-day timestep ensures that there are two separate satellite passes per image mosaic. 

Once imagery is collected for each timestep, values are assigned to individual pixels, pulling first from the image with the lowest amount of cloud cover. If there are masked pixels in this image, pixel values are determined for these locations using the image with the next lowest amount of cloud cover; this process repeats until either all the images available at a timestep are cycled through or each pixel in the 10-day mosaic is filled with valid, non-clouded values. For mosaics that retain invalid pixels due to persistent cloud cover across the timestep (often during the rainy season in Ethiopia, which stretches from June to September), pixel values are assigned via temporal interpolation: Each invalid pixel is given a linearly interpolated value based on the nearest preceding and subsequent image mosaic with a non-clouded value for that pixel.

Image mosaicking is performed bandwise. All 10m and 20m Sentinel-2 bands are extracted (10 bands in total); the 60m coastal aerosol and water vapor bands are ignored, as these bands contain atmospheric information not relevant for the land process monitoring task at hand. The final image processing step involves temporal smoothing of all timeseries using a 3\textsuperscript{rd} order polynomial Savitzky-Golay filter with a window length of 5. 

To assist with temporal interpolation of clouded pixels at the start and end of the specified time period, 82 image mosaics are collected in total – the 72 image mosaics that make up the 2 years of imagery (June 1, 2019 – June 1, 2020, and June 1, 2020 – June 1, 2021), plus 5 additional timesteps before and after the beginning and end of full time period. After interpolation and smoothing, these additional image mosaics are discarded to leave cloud-free, smoothed, 10 band Sentinel-2 imagery for only the desired 72 timesteps. The imagery is then split into annual temporal stacks, with all training and inference done on a single year’s 36 timesteps of imagery.

Table \ref{tab:interp_statistics} presents statistics on the amount of interpolation necessary to construct 10-day S2 timeseries for all of Tigray and Amhara in 2020 and 2021. Of the 18.94B pixel values in the annual EVI timeseries for Tigray, 15.8\% are interpolated in 2020 and 18.3\% are interpolated in 2021. In Amhara, more pixel values require interpolation due to more cloud cover: Of the 54.85B pixel values in the annual Amhara EVI timeseries, 29.6\% are interpolated in 2020 and 25.2\% are interpolated in 2021. In  Tigray and Amhara, more that 95\% of pixel interpolation in both years is a result of the 10\% cloud cover threshold filtering out all imagery collected at a given timestep; the remaining <5\% of interpolations occur for specific pixels deemed as clouds per the ESA Sen2Cor algorithm. 

Intuitively, more pixels are interpolated during the Ethiopian Highlands' rainy season (Jun. 1 -- Oct. 1) than during the dry season (Dec. 1 -- Apr. 1), per Table \ref{tab:interp_statistics}. During the rainy months in Tigray, 32.7\% of pixels are interpolated in 2020 and 44.7\% of pixels are interpolated in 2021; for these same months in Amhara, 54.2\% and 54.4\% of pixels are interpolated in 2020 and 2021, respectively. While these fractions of interpolated pixels appear high, the authors note that EVI timeseries values during the rainy season are not highly predictive of dry season irrigation presence, and as such will have limited impact on classifier behavior. Instead, rainy season EVI values are used to inform the general characteristics of a year's vegetation timeseries, in particular to identify evergreen or barren land cover per the pixel admissibility criteria (see main text, Section 3.3). In contrast, less than 9\% of dry season pixels in Tigray and Amhara are interpolated in both 2020 and 2021.

\begin{table}[h]
 \caption{Sentinel-2 imagery timeseries interpolation statistics.}
  \centering
  \begin{tabular}{lcccc}
    \toprule
     & \Centerstack{Total Fraction\\of Pixels\\Interpolated} &  \Centerstack{Fraction of Pixels\\Interpolated Due to No\\ Imagery at Timestep} & \Centerstack{Fraction of Pixels\\Interpolated in Rainy\\Season (Jun. 1 -- Oct. 1)} & \Centerstack{Fraction of Pixels\\Interpolated in Dry\\Season (Dec. 1 -- Apr. 1)}  \\ \midrule \midrule
     Tigray, 2020 & 0.158 & 0.954 & 0.327 & 0.064 \\ \midrule 
     Tigray, 2021 & 0.183 & 0.976 & 0.447 & 0.010 \\ \midrule 
     Amhara, 2020 & 0.296 & 0.972 & 0.542 & 0.088 \\ \midrule 
     Amhara, 2021 & 0.252 & 0.972 & 0.544 & 0.017 \\ 
    \bottomrule
  \end{tabular}
  \label{tab:interp_statistics}
\end{table}

\subsection{Ground collection}

The ground collection survey was conducted during the months of March and April, 2021. Enumerators collected labels across an area north and east of Lake Tana (referred to as “Tana”; see Figure 1 in the main text) in a process that involved traveling to individual plots of lands, collecting four coordinate points corresponding to the corners of the plot, and specifying whether irrigation was present on the plot during the visit. The ground collection survey team collected 2002 polygons in Tana: 1500 were labeled non-irrigated and 502 were labeled irrigated. In total, these polygons cover 1867 Ha, 78\% of which was designated as non-irrigated.

\subsection{Labeled data accounting}

Tables \ref{tab:num_labeled_polygons} and \ref{tab:num_labeled_samples} present summaries of the number of labeled polygons and the number of labeled samples used in this paper's analysis. 

\begin{table}[h]
 \caption{Summary of labeled polygons, split by region and model training configuration. GC stands for ground collection labels; VC stands for visual collection labels.}
  \centering
  \begin{tabular}{lccccccccc}
    \toprule
    & & \multicolumn{8}{c}{Number of Labeled Polygons} \\
    \cmidrule(r){3-10}
    & & \multicolumn{2}{c}{Training} & \multicolumn{2}{c}{Validation} & \multicolumn{2}{c}{Testing} & \multicolumn{2}{c}{Total} \\
    Region & \Centerstack{Type of \\Labels} &  Non-Irrig. & Irrig. & Non-Irrig. & Irrig. & Non-Irrig. & Irrig. & Non-Irrig. & Irrig. \\
    \midrule \midrule
    Tana & GC & 1050 & 351 & 225 & 76 & 225 & 75 & 1500 & 502 \\
    \midrule
    Rift & VC & 12 & 25 & 3 & 6 & 3 & 6 & 18 & 37 \\
    \midrule
    Koga &  VC & 27 & 46 & 6 & 10 & 6 & 10 & 39 & 66 \\
    \midrule
    Kobo & VC & 26 & 28	& 6 & 6 & 6 & 7 & 38 & 41 \\
    \midrule
    Alamata & VC & 17 & 16 & 4 & 4 & 4 & 4 & 25 & 24 \\
    \midrule
    Liben & VC & 24 & 25 & 5 & 5 &	6 & 6 & 35 & 36 \\
    \midrule
    Jiga & VC & 15 & 13 & 4 & 3 & 3 & 3	& 22 & 19 \\
    \midrule
    Motta & VC & 17 & 17 & 4 & 4 & 4 & 4 & 25 & 25 \\
    \midrule \midrule
    Total & GC + VC & 1188 & 521 & 257 & 114 & 257 & 115 & 1702 & 750 \\
  \end{tabular}
  \label{tab:num_labeled_polygons}
\end{table}

\begin{table}[h]
\footnotesize
 \caption{Summary of labeled samples, split by region and model training configuration. GC stands for ground collection labels; VC stands for visual collection labels.}
  \centering
  \begin{tabular}{lccccccccc}
    \toprule
    & & \multicolumn{8}{c}{Number of Labeled Samples} \\
    \cmidrule(r){3-10}
    & & \multicolumn{2}{c}{Training} & \multicolumn{2}{c}{Validation} & \multicolumn{2}{c}{Testing} & \multicolumn{2}{c}{Total} \\
    Region & \Centerstack{Type of \\Labels} &  Non-Irrig. & Irrig. & Non-Irrig. & Irrig. & Non-Irrig. & Irrig. & Non-Irrig. & Irrig. \\
    \midrule \midrule
    Tana & GC & 63,729 & 24,675 & 14,283 & 5089 & 13,910 & 5361 & 91,922 & 35,125 \\
    \midrule
    Rift & VC & 92,157 & 104,682 & 19,149 & 19,269 & 20,378 & 20,286 & 131,684 & 144,237 \\
    \midrule
    Koga & VC & 150,378 & 98,697 & 29,661 & 23,015 & 27,953 & 24,401 & 207,992 & 146,113 \\
    \midrule
    Kobo &  VC & 93,838 & 123,946 & 30,549 & 36,494	& 31,473 & 48,077 &	155,860 & 208,517 \\
    \midrule
    Alamata & VC & 58,310 & 21,176 & 14,356 & 4601 & 11,083	& 6447 & 83,749 & 32,224 \\
    \midrule
    Liben & VC & 132,999 & 113,733 & 26,027 & 31,212 & 35,394 & 21,895 & 194,420 & 166,840 \\
    \midrule
    Jiga & VC & 113,640 & 79,143 & 33,244 & 15,368 & 38,734 & 12,204 & 185,618 & 106,715 \\
    \midrule
    Motta & VC & 94,153 & 47,915 & 34,267 & 11,127 & 27,568 & 9074 & 155,988 & 68,116 \\
    \midrule \midrule
    Total & GC + VC & 799,204 & 613,967 & 201,536 & 146,175 & 206,493 & 147,745 & 1,207,233 & 907,887
  \end{tabular}
  \label{tab:num_labeled_samples}
\end{table}

\subsection{Labeled data exploration}

To better understand the vegetation phenologies contained within this study’s labeled data, the similarities of EVI timeseries of the same class are explored across regions. This process first involves applying a PC transform to all labeled training data. The samples’ dimensionality is then reduced by using only the first 10 dimensions of the transformed data; these first 10 dimensions explain 91\% of the variance contained within the samples’ EVI timeseries.

\begin{table}[h]
 \caption{Pairwise pseudo-1D Kolmogorov-Smirnov (KS) statistics between regions’ non-irrigated training samples. Values with typographical symbols are to be interpreted alongside Table \ref{tab:f1_score_regression}.}
  \centering
  \begin{tabular}{lcccccccc|c}
    \toprule
     & Tana	& Rift & Koga & Kobo & Alamata & Liben & Jiga & Motta & Mean \\ \midrule \midrule
     Tana & 0.00 & 1.13 & 1.85 & 1.00 & 1.25 & 1.69 & 1.50 & 1.33 & 1.39 \\ \midrule
Rift & 1.13 & 0.00 & 1.48 & 0.62 & 0.71 & 1.43 & 0.95 & 0.57 & 0.98 \\ \midrule
Koga & 1.85$\ssymbol{1}$ & 1.48 & 0.00 & 1.48 & 1.33 & 0.41 & 1.02 & 1.26 & 1.26 \\ \midrule
Kobo & 1.00 & 0.62 & 1.48 & 0.00 & 0.76 & 1.45 & 1.16 & 0.91 & 1.05 \\ \midrule
Alamata & 1.25 & 0.71 & 1.33 & 0.76 & 0.00 & 1.33 & 0.97 & 0.72 & 1.01$\ssymbol{2}$ \\ \midrule
Liben & 1.69 & 1.43 & 0.41 & 1.45 & 1.33 & 0.00 & 0.91 & 1.23 & 1.21 \\ \midrule
Jiga & 1.50$\ssymbol{7}$ & 0.95 & 1.02 & 1.16 & 0.97 & 0.91 & 0.00 & 0.51 & 1.00 \\ \midrule
Motta & 1.33 & 0.57 & 1.26 & 0.91 & 0.72 & 1.23 & 0.51 & 0.00 & 0.93$\ssymbol{3}$ \\ \midrule \midrule
 	& & & & & & & & & 1.11 \\
    \bottomrule
  \end{tabular}
  \label{tab:ks_distance_non-irrigated}
\end{table}

\begin{table}[h]
 \caption{Pairwise pseudo-1D Kolmogorov-Smirnov (KS) statistics between regions’ irrigated training samples. Values with typographical symbols are to be interpreted alongside Table \ref{tab:f1_score_regression}.}
  \centering
  \begin{tabular}{lcccccccc|c}
    \toprule
      
 & Tana & Rift & Koga & Kobo & Alamata & Liben & Jiga & Motta & Mean \\ \midrule \midrule
Tana & 0.00 & 1.88 & 2.25 & 1.73 & 1.65 & 1.13 & 1.60 & 1.27 & 1.64 \\ \midrule
Rift & 1.88 & 0.00 & 1.48 & 0.61 & 0.37 & 2.01 & 1.30 & 1.51 & 1.31 \\ \midrule
Koga & 2.25$\ssymbol{1}$ & 1.48 & 0.00 & 1.37 & 1.45 & 2.46 & 1.98 & 2.20 & 1.88 \\ \midrule
Kobo & 1.73 & 0.61 & 1.37 & 0.00 & 0.65 & 1.98 & 1.51 & 1.57 & 1.35 \\ \midrule
Alamata & 1.65 & 0.37 & 1.45 & 0.65 & 0.00 & 1.78 & 1.11 & 1.31 & 1.19$\ssymbol{2}$ \\ \midrule
Liben & 1.13 & 2.01 & 2.46 & 1.98 & 1.78 & 0.00 & 1.41 & 0.98 & 1.68 \\ \midrule
Jiga & 1.60$\ssymbol{7}$ & 1.30 & 1.98 & 1.51 & 1.11 & 1.41 & 0.00 & 0.89 & 1.40 \\ \midrule
Motta & 1.27 & 1.51 & 2.20 & 1.57 & 1.31 & 0.98 & 0.89 & 0.00 & 1.39$\ssymbol{3}$ \\ \midrule  \midrule
& & & & & & & & & 1.48 \\
    \bottomrule
  \end{tabular}
  \label{tab:ks_distance_irrigated}
\end{table}

After dimensionality reduction via the PC transform, the two sample Kolmogorov-Smirnov (KS) test statistic is calculated between sample distributions of the same class across regions. The two-sample KS statistic determines the largest absolute distance between two 1D empirical distributions, and is presented in Eq. \ref{ks}: 

\begin{equation}
D_{KS}= sup_{x}\lvert F_1(x) - F_2(x)\rvert
\label{ks}
\end{equation}

where F\textsubscript{1} and F\textsubscript{2} are the two empirical distribution functions of 1D variable \textit{x}, and \textit{sup} is the supremum function. The KS statistic is assessed for two reasons: 1) the statistic depends on no assumptions about the underlying data distributions; and 2) the statistic has been adapted for multivariate distributions via the pseudo-1D KS metric \citep{Hagen2020}. In this adaptation, the pseudo-1D KS metric, $D_{KS,P1D}$, is the Euclidean KS statistic calculated between successive orthogonal dimensions of two multivariate distributions: 

\begin{equation}
D_{KS,P1D}= \sqrt{(D_{KS,1})^2 + (D_{KS,2})^2 + \cdots + (D_{KS,n})^2} 
\label{d_ks_p1d}
\end{equation}

where

\begin{equation}
D_{KS,n}= sup_{y_n}\lvert F_1(y_n) - F_2(y_n)\rvert
\label{d_ks_n}
\end{equation}

Here, Eq. \ref{d_ks_n} represents the KS statistic between the empirical distribution functions F\textsubscript{1} and F\textsubscript{2} of the \textit{n\textsuperscript{th}} dimension of multivariate variable \textit{y}. As only the first 10 PC dimensions of the transformed data are used, \textit{n} ranges between 1 and 10. 

Table \ref{tab:ks_distance_non-irrigated} presents pairwise pseudo-1D KS statistics between regions’ non-irrigated samples; Table \ref{tab:ks_distance_irrigated} presents pairwise pseudo-1D KS distances between regions’ irrigated samples. Here, the relative statistics between distributions are compared, as the absolute statistics cannot be interpreted in a physically meaningful way. The cells with typographical marks in the two tables indicate statistics to be interpreted with the results in Table \ref{tab:f1_score_regression}, discussed alongside that table. Tables \ref{tab:ks_distance_non-irrigated} and \ref{tab:ks_distance_irrigated} show that the relative pairwise statistic between regional distributions is larger among the irrigated sample sets, indicating that irrigated samples are more dissimilar across regions compared to the non-irrigated samples. This takeaway reflects the varying nature of irrigation practices across Ethiopia – irrigation can occur at different parts of the dry season for a variety of different crops. In contrast, the phenologies of non-irrigated cropland must mirror Ethiopia’s primary rains, which are consistent in time for the regions included in this analysis.

\begin{figure}[h]
\begin{center}
\includegraphics[width=\textwidth]{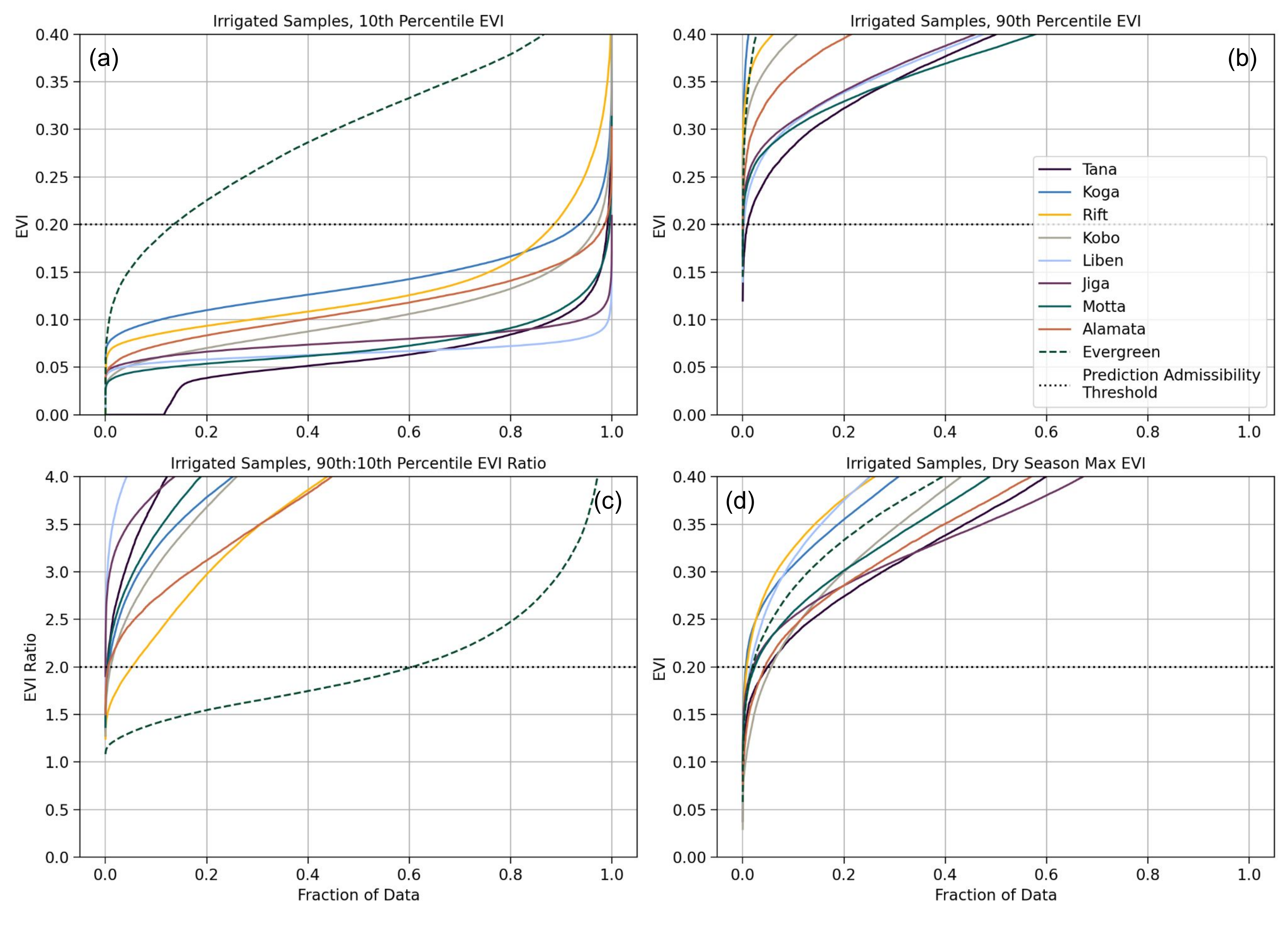}
\end{center}
\caption{Cumulative distribution functions (CDFs) for the (a) 10\textsuperscript{th} and (b) 90\textsuperscript{th} EVI timeseries percentiles; (c) the 90\textsuperscript{th}:10\textsuperscript{th} EVI timeseries percentile ratio; and (d) the maximum EVI value during the dry season (December 1\textsuperscript{st}, 2020, to April 1\textsuperscript{th}).}
\label{pixel_cdfs}
\end{figure}

\subsection{Determining the prediction admissibility criteria}

The prediction admissibility criteria presented in the Methodology of the main text (see Table 1) are informed by the cumulative distribution functions (CDFs) of the collected samples’ EVI timeseries. By imposing admissibility criteria that closely match the distribution of the samples' EVI timeseries, heuristics are devised to exclude many pixels not relevant to the non-irrigated/irrigated cropland prediction methodology. Figure \ref{pixel_cdfs} presents CDFs for the 10\textsuperscript{th} and 90\textsuperscript{th} EVI timeseries percentiles, the 90\textsuperscript{th}:10\textsuperscript{th} EVI timeseries percentile ratio, and the maximum EVI value during the dry season. Values are presented for all regions’ irrigated samples only, along with a set of pixel timeseries over evergreen areas.

Figure \ref{pixel_cdfs}(a) shows that a maximum of 0.2 for the 10\textsuperscript{th} percentile of the EVI timeseries is achieved by nearly all irrigated samples, and how this ceiling filters out 85\% of all evergreen samples. Similarly, a minimum 90\textsuperscript{th}:10\textsuperscript{th} percentile EVI ratio of 2 is satisfied by nearly all irrigated samples and excludes 60\% of evergreen samples (Figure \ref{pixel_cdfs}(c)). While no EVI timeseries for barren or non-vegetated areas are shown in this figure, the criteria specifying a 90\textsuperscript{th} percentile EVI value above 0.2 and a dry season max EVI value above 0.2 are met by the vast majority of irrigated samples (Figure \ref{pixel_cdfs}(b,d)), and would filter out many of these non-cropped pixels.

\section{Supplementary Results}
\subsection{Evaluating the effects of randomly shifted input timeseries with a Gradient-Class Activation Map}

\begin{figure}[h]
\begin{center}
\includegraphics[width=\textwidth]{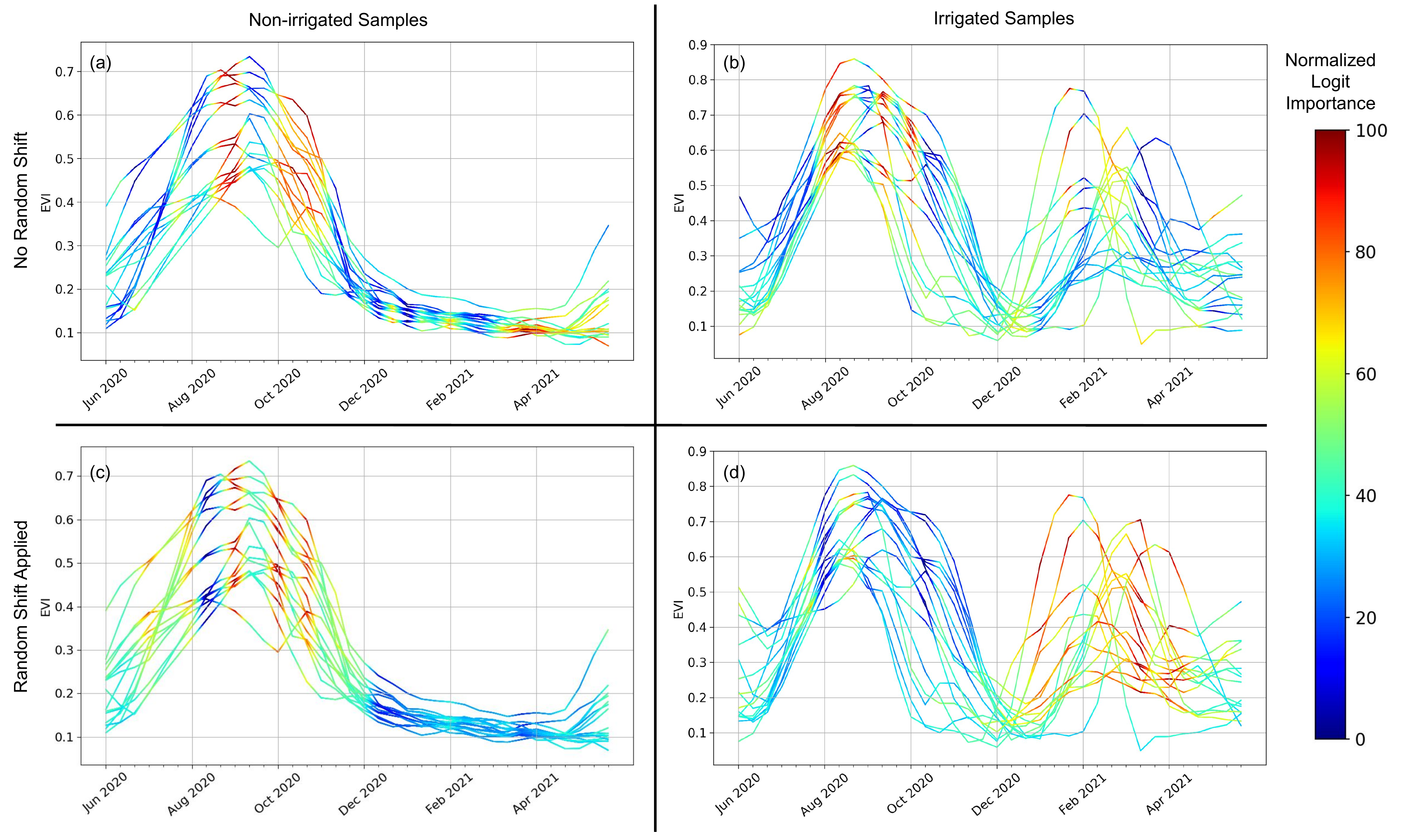}
\end{center}
\caption{Modified Grad-CAM timestep importances for 16 randomly selected non-irrigated and irrigated enhanced vegetation index (EVI) timeseries from Koga, before and after the timeseries shift is applied.}
\label{gradcam}
\end{figure}

Implementing a modified Gradient-Class Activation Map (Grad-CAM) for visual prediction explanation provides further evidence for improved prediction robustness from randomly shifting input EVI timeseries. A Grad-CAM uses the gradients flowing into the final layer of a neural network to produce a localization map highlighting important portions of the input for predicting a concept. Originally developed for images in \citet{Selvaraju2020}, this technique can be applied analogously to timeseries. To do so, a transformer-based classifier model with its 32-node penultimate dense layer removed is trained on all VC regions’ training datasets; by removing this fully connected layer, the importance of each timestep input for prediction can be visualized, as there is no longer a layer obscuring the gradient flow into the final prediction nodes. Figure \ref{gradcam} displays the normalized timestep prediction importances for 16 randomly selected non-irrigated and irrigated EVI timeseries from the Koga region. Results are presented for two models: the first trained without randomly shifting input timeseries, and the second trained with the random shift applied. 

Figure \ref{gradcam} demonstrates that the input timesteps most important for prediction – displayed in red per the Normalized Logit Importance colorbar – are more continuous and better reflect a common understanding of what portions of a phenology should be predictive once the model is trained on randomly shifted input timeseries. For the non-irrigated samples, the model trained on the randomly shifted timeseries identifies a larger portion of the timesteps during the rainy season as highly predictive (Figure \ref{gradcam}(c)); this model also correctly identifies vegetation growth during dry season timesteps as important for identifying irrigated samples (Figure \ref{gradcam}(d)). In comparison, the model trained on the non-shifted timeseries identifies scattered timesteps as predictive for both non-irrigated and irrigated samples (Figure \ref{gradcam}(a,b)); it does not emphasize dry season vegetation growth as predictive of irrigation presence (Figure \ref{gradcam}(b)). Instead, this model learns to identify isolated, non-intuitive timesteps, and as a consequence is more likely to misclassify input timeseries that differ slightly from those in the training data. 

\subsection{Ablation study: Limiting polygons used during training}

To understand the impact of the amount of training data on model performance, an ablation study is conducted where the fraction of labeled polygons included in each region’s training dataset is varied between 0.15 and 0.85; the complementary fraction of each region’s polygons comprises the test dataset. For each fraction of training polygons, the CatBoost model architecture is trained on all combinations of all 7 visual collection (VC) regions’ training datasets; performance is assessed on the withheld VC regions in a process identical to the one described for main text Figures 5 and 6. All models are trained on randomly shifted EVI timeseries. 

\begin{figure}[h]
\begin{center}
\includegraphics[width=11cm]{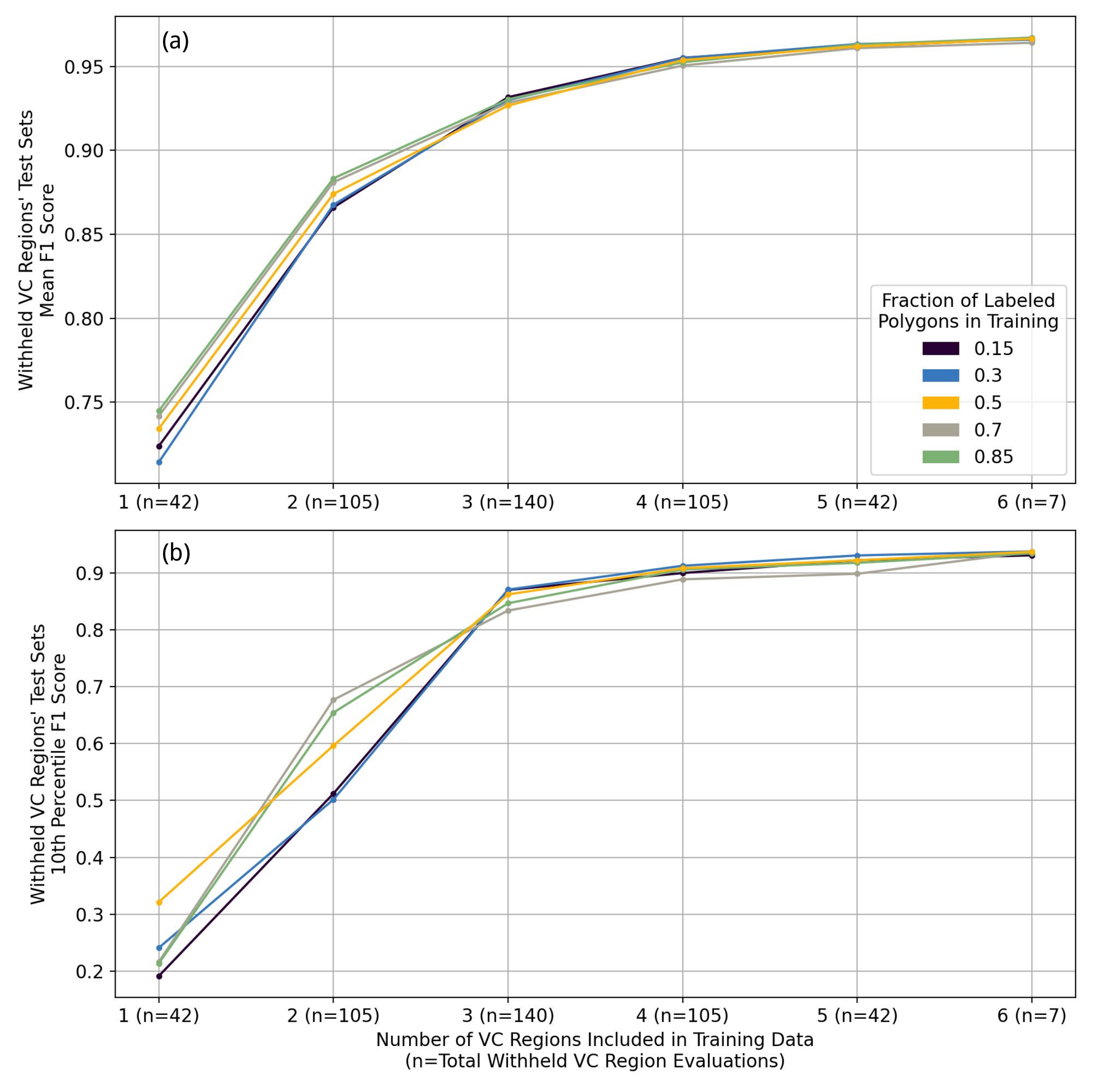}
\end{center}
\caption{Withheld region test dataset performance for different fractions of labeled polygons included in the training datasets; the complementary fractions of labeled polygons constitute the test datasets. Predictions are made using a CatBoost model architecture. (a) presents mean F1 score over the withheld regions; (b) presents the 10th percentile F1 score over the withheld regions.}
\label{ablation_fig}
\end{figure}

Figure \ref{ablation_fig} presents the results of this ablation study, in which minimal impact is observed when varying the fraction of polygons included in each region’s training dataset between 0.15 and 0.85.  On average, withheld region F1 score decreases by approximately 0.05 as the fraction of training polygons drops from 0.85 to 0.15 when 1 VC region is included in the training data; this gap shrinks as additional VC regions are incorporated during training, becoming negligible for all models trained on 3 or more VC regions’ data. A larger performance delta among the 10th percentile of withheld regions’ F1 scores exists when fewer than 3 VC regions are included during training; similar to the average performance metrics, this gap collapses when the classification model is trained on labeled data from 3 or more VC regions. 

Figure \ref{ablation_fig} demonstrates that the irrigation prediction models are robust even when limiting the fraction of polygons included in training datasets to 15\% of the total. Here, the inclusion of labeled data from multiple regions and randomly shifting the EVI timeseries inputs introduces enough variance to the classification model during training so that performance over regions unseen by the classifier remains high. 

\subsection{Comparison of predictions across model architectures}

To ensure that different irrigation detection architectures converge on similar decision boundaries, the alignment of predictions across the transformer-based and CatBoost models is investigated. Here, both architectures are training on the randomly shifted EVI timeseries of all 7 VC regions’ training datasets; predictions are then made over all labeled samples. Table \ref{tab:transformer_catboost_comparison} presents the alignment of these predictions, showing a high degree of prediction similarity: An average regional prediction alignment of 98.9\% is calculated. The close alignment of predictions made by both these models expands the basis for the solution set. 

\begin{table}[h]
 \caption{Comparison between Transformer and CatBoost model predictions for models trained on all 7 visual collection (VC) regions’ training datasets.}
  \centering
  \begin{tabular}{lcccccc}
    \toprule
    Region & \makecell{Type of\\Labels}	& \multicolumn{2}{c}{\makecell{Num. Aligned Sample\\Predictions}} &  \multicolumn{2}{c}{\makecell{Num. Misaligned Sample\\Predictions}}	& \makecell{Fraction of\\aligned predictions} \\ \midrule
     & &  Non-Irrig. (0) & Irrig. (1) & \makecell{Transformer: 0\\CatBoost: 1} & \makecell{Transformer: 1\\CatBoost: 0} & \\ \midrule \midrule
 Tana & GC	    & 88,587 & 35,035 & 728 & 2690 & 0.973 \\ \midrule
Rift & VC	    & 130,399 & 142,114 & 2367 & 1024 & 0.988 \\ \midrule
Koga & VC	    & 207,536 & 144,284 & 1401 & 827 & 0.994 \\ \midrule
Kobo & VC	    & 156,946 & 203,510 & 2313 & 583 & 0.992 \\ \midrule
Alamata & VC    & 84,024 & 31,978 & 615 & 251 & 0.993 \\ \midrule
Liben & VC      & 193,428 & 165,933 & 884 & 971 & 0.995 \\ \midrule
Jiga & VC	    & 184,116 & 106,384 & 955 & 833 & 0.994 \\ \midrule
Motta & VC	    & 155,056 & 65,956 & 1689 & 1363 & 0.986 \\
    \bottomrule
  \end{tabular}
  \label{tab:transformer_catboost_comparison}
\end{table}

\subsection{Determining the relative importance of each visual collection region}

Through a pair of ordinary least-squares (OLS) regressions, the contribution of each VC region to target region performance can be assessed. Table \ref{tab:f1_score_regression} presents OLS regression coefficients and P-values on target region F1 scores for the 7 VC regions used during training, where the F1 scores are collected over all withheld regions for all transformer classifier models presented in Figure 6 of the main text. In interpreting the regression results, variables with P-values above 0.05 are considered not statistically significant. 

Table \ref{tab:f1_score_regression} shows that training data from the regions of Alamata or Kobo have the largest impact on Tana test dataset performance, increasing F1 score on average by 0.032 or 0.024, respectively. The non-statistically significant contributions of Koga and Jiga’s training data to Tana test dataset performance are highlighted, shown by the values marked with $\ssymbol{1}$ and $\ssymbol{7}$. Comparing these non-statistically significant results with the relevant cells in Tables \ref{tab:ks_distance_non-irrigated} and \ref{tab:ks_distance_irrigated} – also marked with $\ssymbol{1}$ and $\ssymbol{7}$ – reveals that non-irrigated and irrigated samples from both Koga and Jiga are more dissimilar from Tana labeled samples compared to the regional average, determined by the KS distance between the regions’ data. 

\begin{table}[ht]
 \caption{Ordinary least squares regression on withheld target region F1 scores. F1 scores are collected over all withheld regions for all transformer classifier models presented in Figure 6 of the main text. Values with typographical symbols are to be interpreted alongside Tables \ref{tab:ks_distance_non-irrigated} and \ref{tab:ks_distance_irrigated}.}
  \centering
  \begin{tabular}{lcccc}
    \toprule
    & \multicolumn{2}{c}{\makecell{Tana\\(R\textsuperscript{2}=0.317, n=127)}} & \multicolumn{2}{c}{\makecell{Withheld VC Regions\\(R\textsuperscript{2}=0.18, n=441)}} \\ \midrule
    Source VC Region & Coefficient & P-value & Coefficient & P-value \\ \midrule \midrule
    Rift & 0.018 & 0.008 & -0.012 & 0.327 \\ \midrule
    Koga & 0.013$\ssymbol{1}$ & 0.056$\ssymbol{1}$ & 0.016 & 0.172 \\ \midrule
    Kobo & 0.024 & 0.001 & 0.029 & 0.016 \\ \midrule
    Alamata & 0.032 & 0.000 & 0.071$\ssymbol{2}$ & 0.000$\ssymbol{2}$ \\ \midrule
    Liben & -0.011 & 0.122 & 0.028 & 0.019 \\ \midrule
    Jiga & 0.013$\ssymbol{7}$ & 0.063$\ssymbol{7}$ & 0.047 & 0.000 \\ \midrule
    Motta & 0.016 & 0.024 & 0.080$\ssymbol{3}$ & 0.000$\ssymbol{3}$ \\ 
    \bottomrule
  \end{tabular}
  \label{tab:f1_score_regression}
\end{table}

Table \ref{tab:f1_score_regression} also shows that the inclusion of labeled samples from Motta and Alamata during training causes the largest increase in F1 score over the withheld VC regions; these increases, shown by values marked with $\ssymbol{2}$ and $\ssymbol{3}$, amount to 0.08 and 0.071, respectively.  Again, comparing these data points to the KS distances marked with $\ssymbol{2}$ and $\ssymbol{3}$ in Tables \ref{tab:ks_distance_non-irrigated} and \ref{tab:ks_distance_irrigated} demonstrates that non-irrigated and irrigated samples from Motta and Alamata are more similar to withheld VC regions’ labeled data on average, as compared to samples from other VC regions.
 
Taken together, the results from Tables \ref{tab:ks_distance_non-irrigated}, \ref{tab:ks_distance_irrigated}, and \ref{tab:f1_score_regression} yield the intuitive finding that labeled samples more similar to those in a target region have a greater positive impact on performance, while more dissimilar labeled samples have a weaker effect on performance. 

\subsection{Independently labeled polygons for performance assessment at inference}

\begin{figure}[h]
\begin{center}
\includegraphics[width=\textwidth]{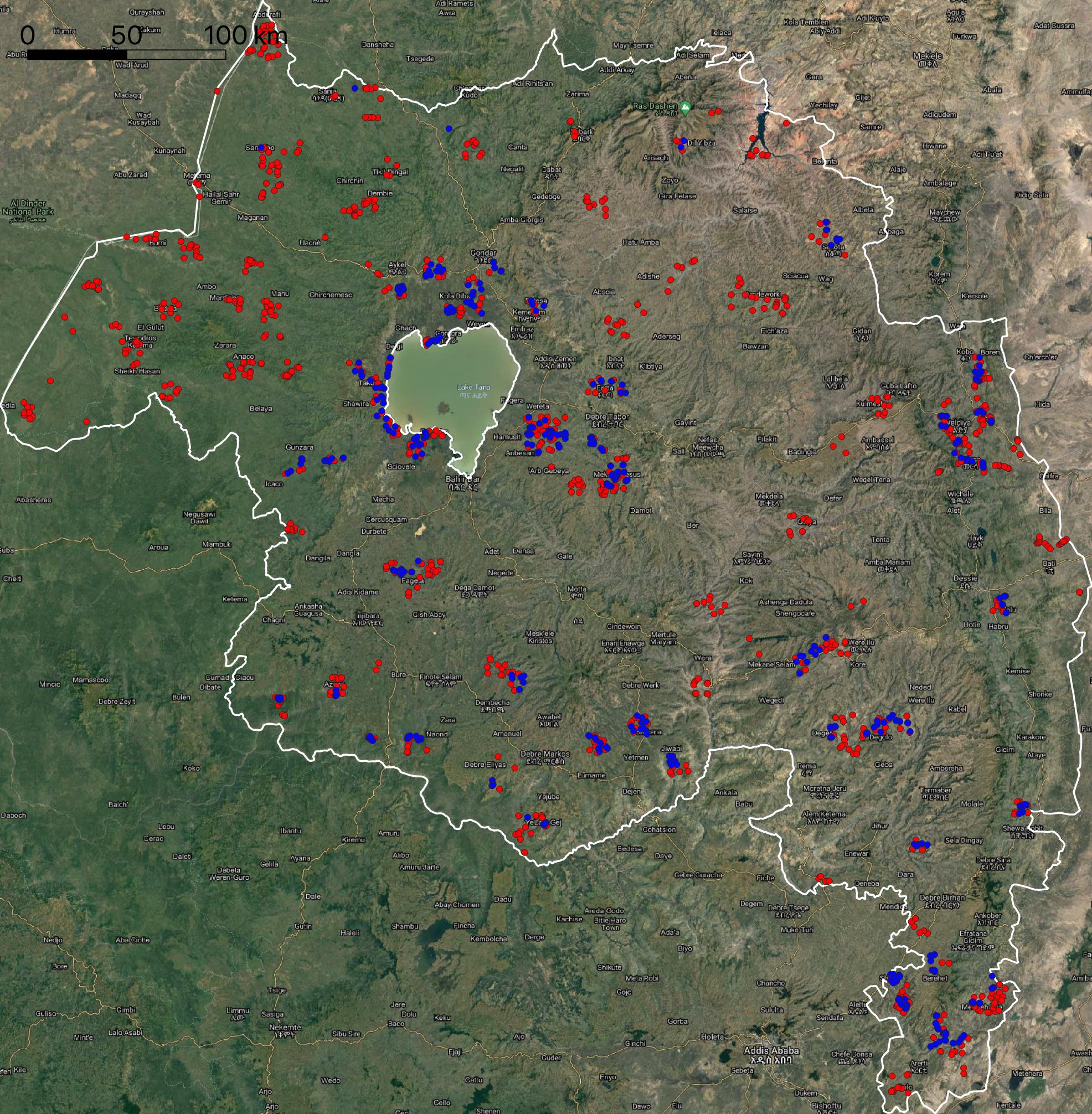}
\end{center}
\caption{Locations of independently labeled polygons for additional model performance assessment. The centroids of non-irrigated polygons are shown in red, 1082 in total; the centroids of irrigated polygons are shown in blue, 519 in total. These polygons produce 361,451 non-irrigated samples and 48,465 irrigated timeseries samples.}
\label{inference_locations}
\end{figure}

Figure \ref{inference_locations} presents the centroids of all polygons collected by additional enumerators for prediction performance assessment at model inference. Non-irrigated polygon locations are shown in red and irrigated polygons locations are shown in blue. 

\bibliographystyle{abbrvnat}
\bibliography{references}  

\section*{Appendix A}

Appendix A presents labeled samples before and after cluster cleaning for all label collection regions except Koga, which is presented in Figure 3 of the main text.

\begin{figure}[h]
\begin{center}
\includegraphics[width=\textwidth]{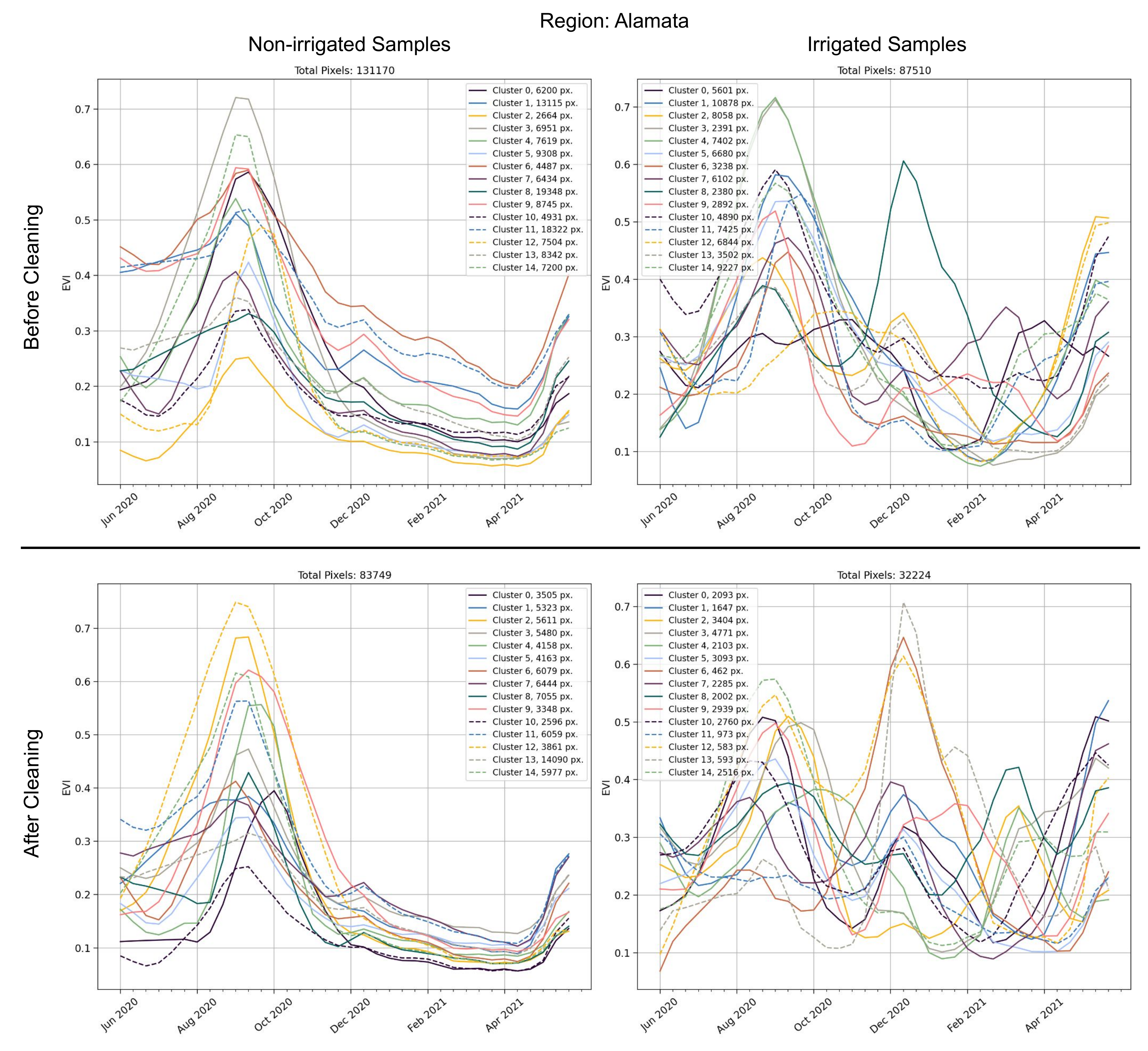}
\end{center}
\caption{Clustered enhanced vegetation index (EVI) timeseries before and after cluster cleaning for the Alamata region. After cleaning, all non-irrigated clusters display a single vegetation peak aligned with the main rainy season, and the irrigated clusters all display a vegetation cycle during the dry season.}\label{alamata_cleaning_labels}
\end{figure}

\begin{figure}[ht]
\begin{center}
\includegraphics[width=\textwidth]{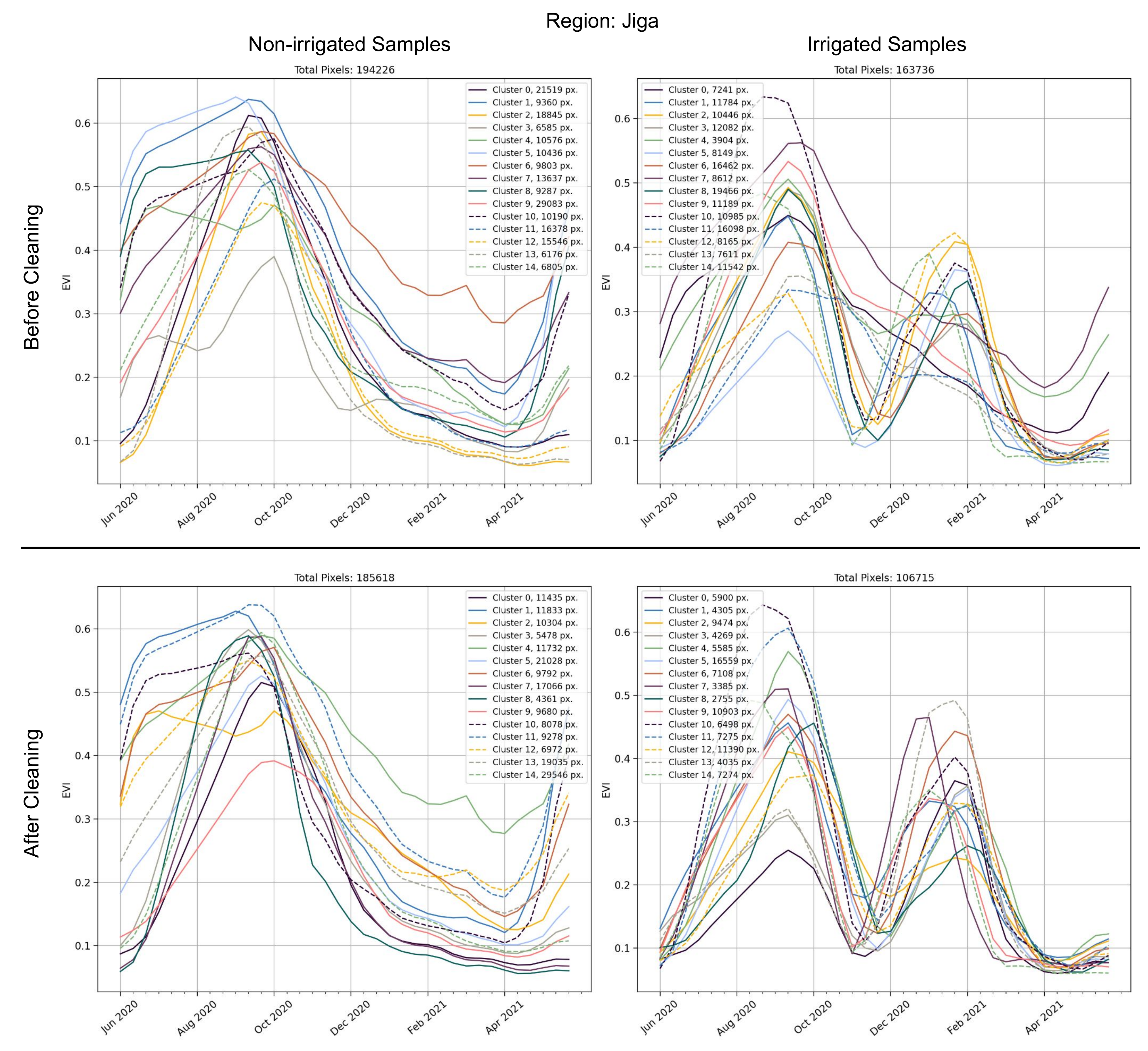}
\end{center}
\caption{Clustered enhanced vegetation index (EVI) timeseries before and after cluster cleaning for the Jiga region. After cleaning, all non-irrigated clusters display a single vegetation peak aligned with the main rainy season, and the irrigated clusters all display a vegetation cycle during the dry season.}\label{jiga_cleaning_labels}
\end{figure}

\begin{figure}[ht]
\begin{center}
\includegraphics[width=\textwidth]{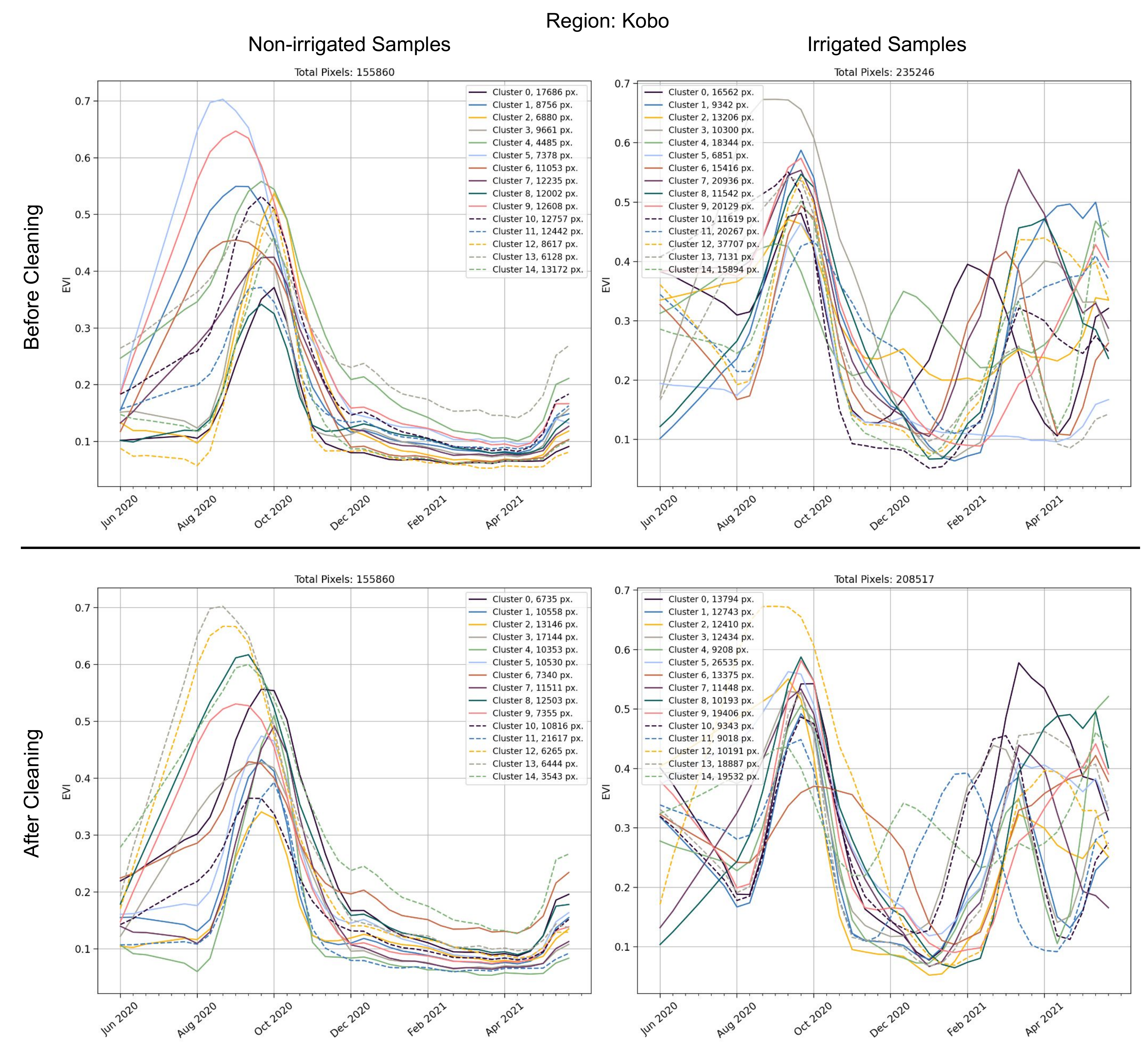}
\end{center}
\caption{Clustered enhanced vegetation index (EVI) timeseries before and after cluster cleaning for the Kobo region. After cleaning, all non-irrigated clusters display a single vegetation peak aligned with the main rainy season, and the irrigated clusters all display a vegetation cycle during the dry season.}\label{kobo_cleaning_labels}
\end{figure}

\begin{figure}[ht]
\begin{center}
\includegraphics[width=\textwidth]{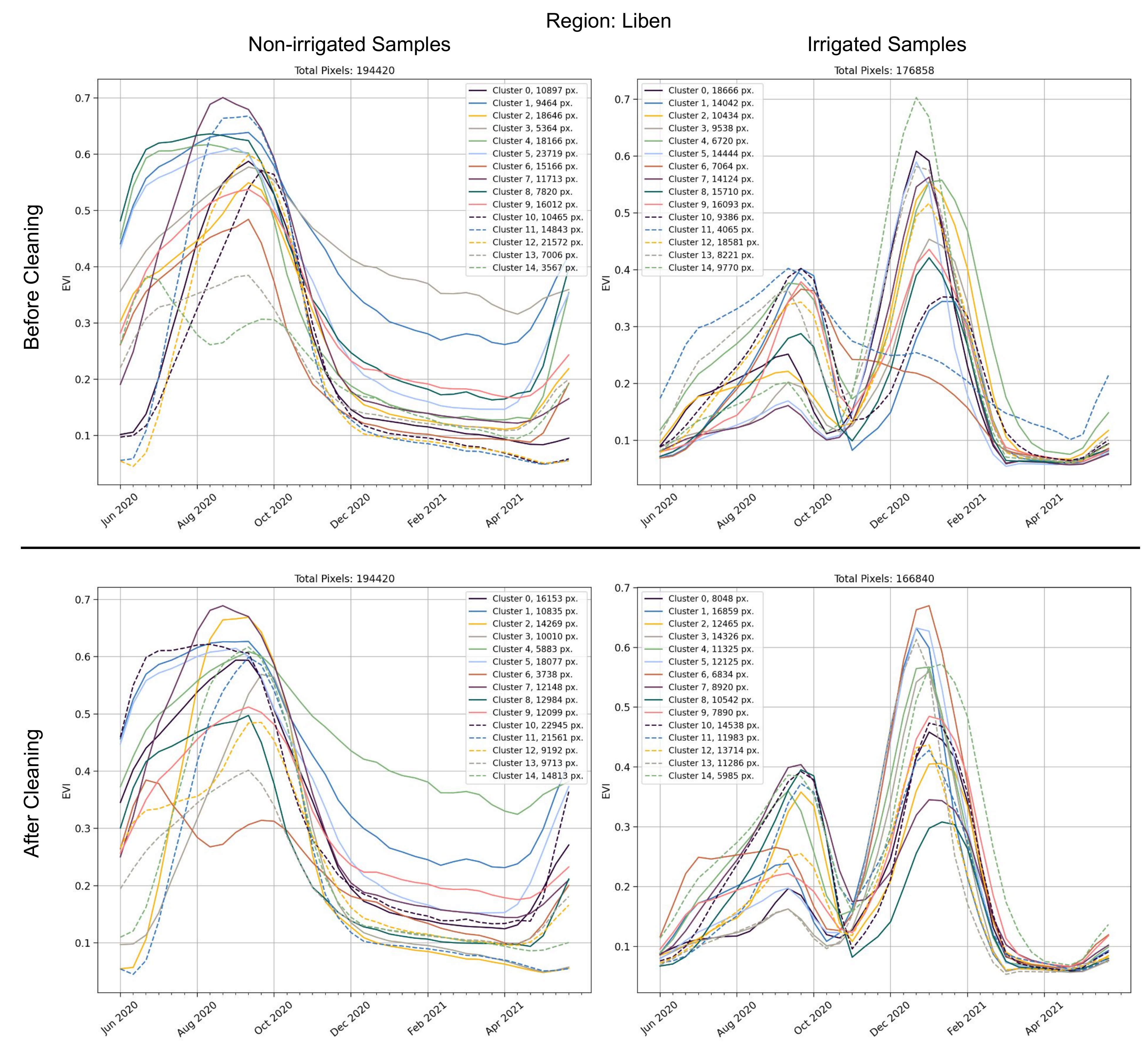}
\end{center}
\caption{Clustered enhanced vegetation index (EVI) timeseries before and after cluster cleaning for the Liben region. After cleaning, all non-irrigated clusters display a single vegetation peak aligned with the main rainy season, and the irrigated clusters all display a vegetation cycle during the dry season.}\label{liben_cleaning_labels}
\end{figure}

\begin{figure}[ht]
\begin{center}
\includegraphics[width=\textwidth]{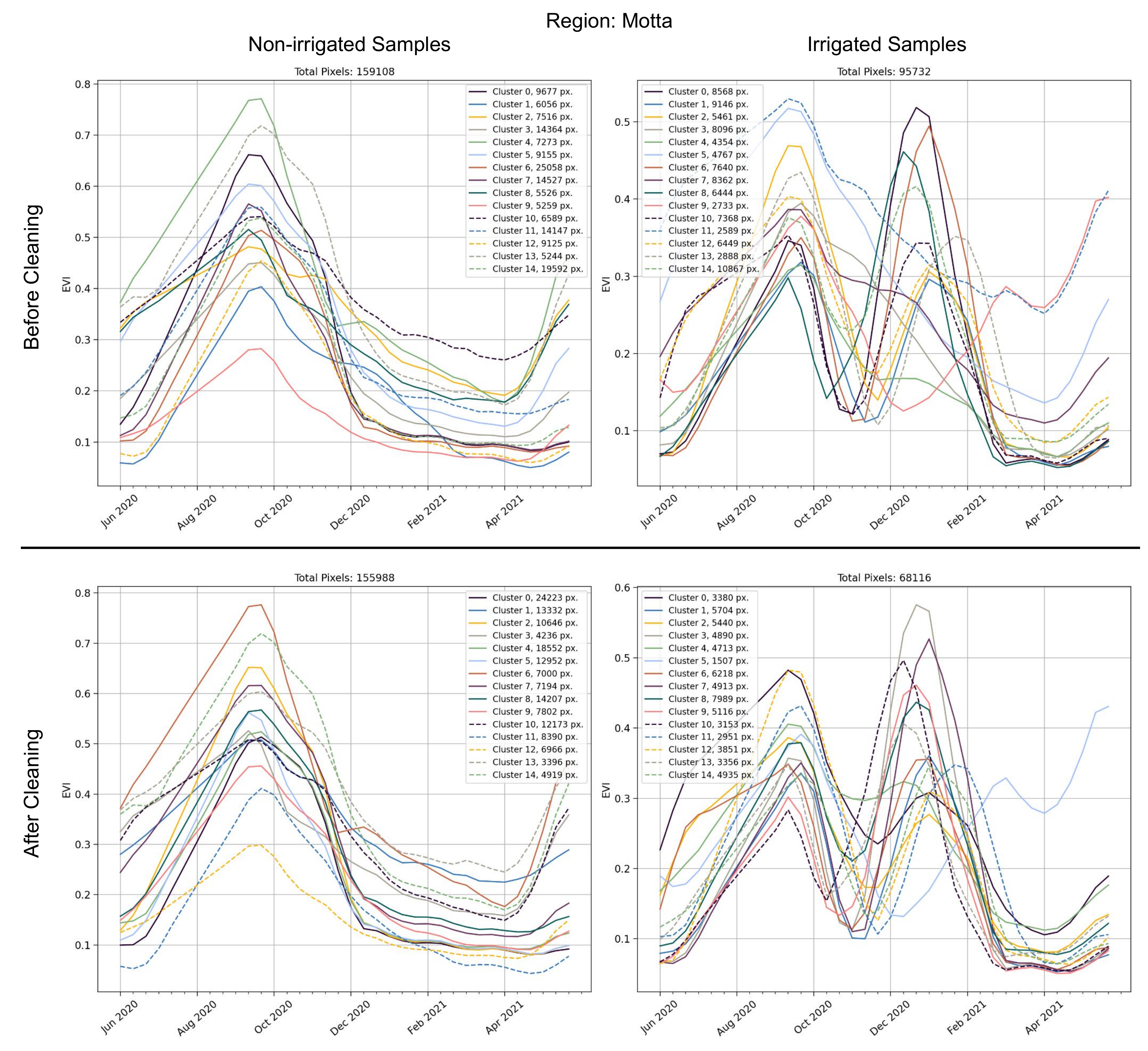}
\end{center}
\caption{Clustered enhanced vegetation index (EVI) timeseries before and after cluster cleaning for the Motta region. After cleaning, all non-irrigated clusters display a single vegetation peak aligned with the main rainy season, and the irrigated clusters all display a vegetation cycle during the dry season.}\label{motta_cleaning_labels}
\end{figure}

\begin{figure}[ht]
\begin{center}
\includegraphics[width=\textwidth]{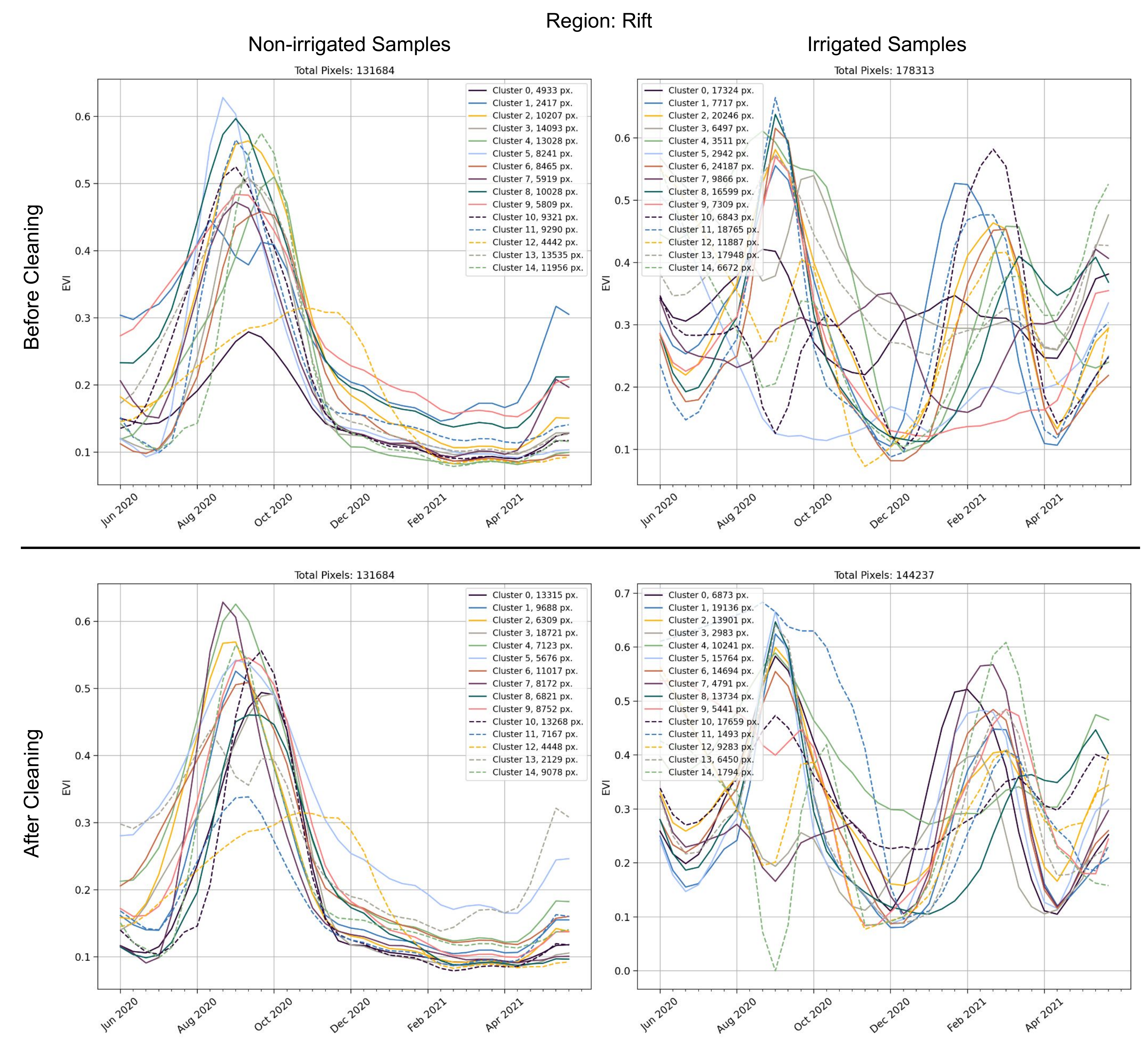}
\end{center}
\caption{Clustered enhanced vegetation index (EVI) timeseries before and after cluster cleaning for the Rift region. After cleaning, all non-irrigated clusters display a single vegetation peak aligned with the main rainy season, and the irrigated clusters all display a vegetation cycle during the dry season.}\label{rift_cleaning_labels}
\end{figure}

\begin{figure}[ht]
\begin{center}
\includegraphics[width=\textwidth]{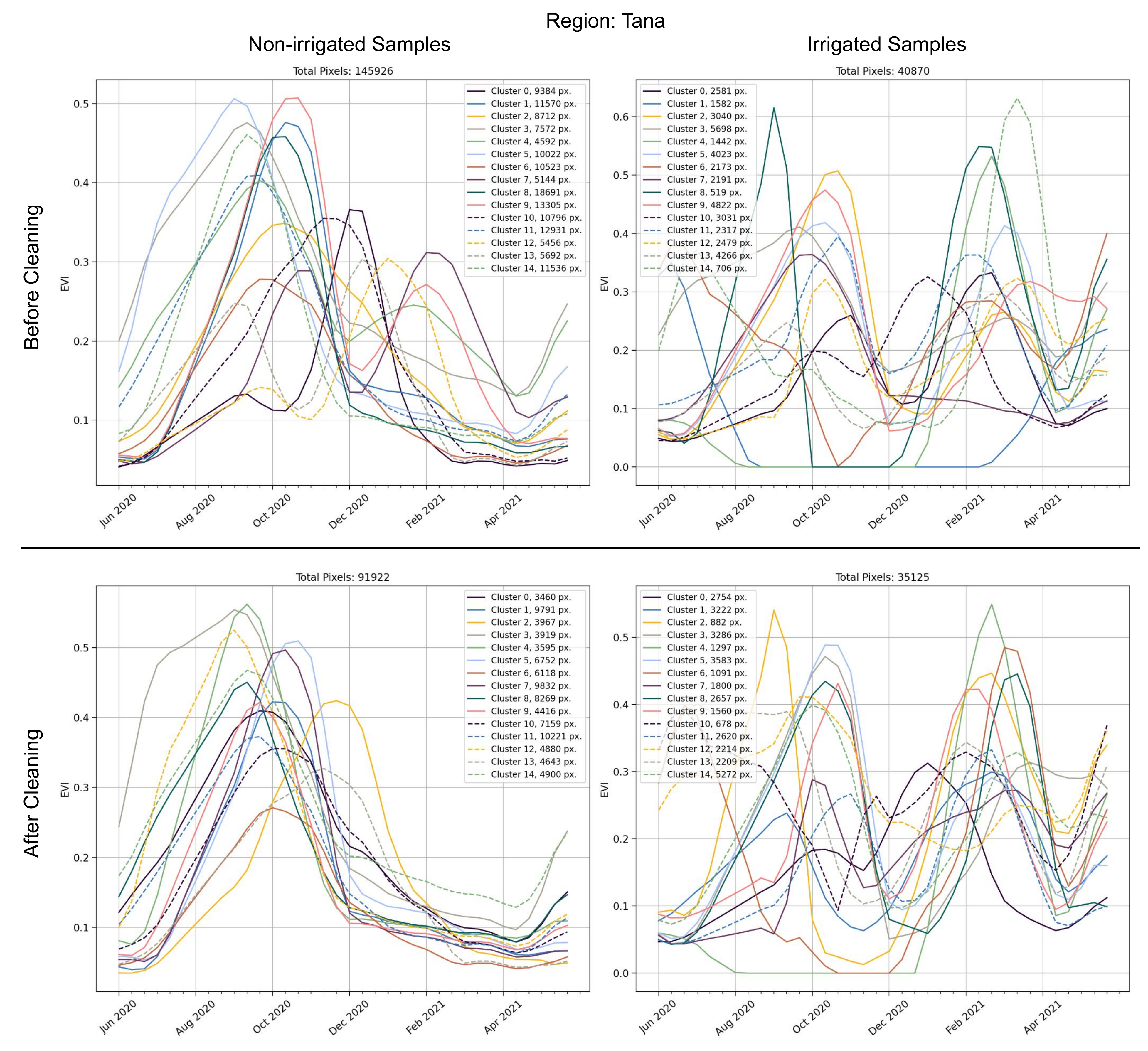}
\end{center}
\caption{Clustered enhanced vegetation index (EVI) timeseries before and after cluster cleaning for the Tana region. After cleaning, all non-irrigated clusters display a single vegetation peak aligned with the main rainy season, and the irrigated clusters all display a vegetation cycle during the dry season.}\label{tana_cleaning_labels}
\end{figure}




